\pdfoutput=1
\documentclass[letterpaper]{article} 
\usepackage{aaai23} 
\usepackage{times}  
\usepackage{helvet}  
\usepackage{courier}  
\usepackage[hyphens]{url}  
\usepackage{graphicx} 
\usepackage{natbib}  
\usepackage{caption} 
\usepackage{algorithm}
\usepackage{algorithmic}

\usepackage{caption}
\usepackage{subcaption}
\usepackage{amsmath}

\usepackage{newfloat}
\usepackage{listings}
\DeclareCaptionStyle{ruled}{labelfont=normalfont,labelsep=colon,strut=off} 
\floatstyle{ruled}
\newfloat{listing}{tb}{lst}{}
\floatname{listing}{Listing}
\nocopyright
\title{Application of deep and reinforcement learning to boundary control problems}
\author {
    Zenin Easa Panthakkalakath,\textsuperscript{\rm 1}
    Juraj Kardoš, \textsuperscript{\rm 1}
    Olaf Schenk \textsuperscript{\rm 1}
}
\affiliations {
    \textsuperscript{\rm 1} Università della Svizzera italiana\\
    zenin.easa.panthakkalakath@usi.ch, juraj.kardos@usi.ch, olaf.schenk@usi.ch
}

\usepackage{bibentry}

\newcommand{\citesubject}[1]{\citeauthor{#1} \cite{#1}}

\begin{document}

\maketitle

\begin{abstract}
    The boundary control problem is a non-convex optimization and control problem in many scientific domains, including fluid mechanics, structural engineering, and heat transfer optimization. The aim is to find the optimal values for the domain boundaries such that the enclosed domain adhering to the governing equations attains the desired state values. Traditionally, non-linear optimization methods, such as the Interior-Point method (IPM), are used to solve such problems.

    This project explores the possibilities of using deep learning and reinforcement learning to solve boundary control problems. We adhere to the framework of iterative optimization strategies, employing a spatial neural network to construct well-informed initial guesses, and a spatio-temporal neural network learns the iterative optimization algorithm using policy gradients. Synthetic data, generated from the problems formulated in the literature, is used for training, testing and validation. The numerical experiments indicate that the proposed method can rival the speed and accuracy of existing solvers. In our preliminary results, the network attains costs lower than IPOPT, a state-of-the-art non-linear IPM, in 51\% cases. The overall number of floating point operations in the proposed method is similar to that of IPOPT. Additionally, the informed initial guess method and the learned momentum-like behaviour in the optimizer method are incorporated to avoid convergence to local minima.
\end{abstract}

\section{Introduction}

Optimal control problems arise in a plethora of applications, many requiring accurate and fast solution methods. Using traditional solution methods is time-consuming; some simulations often take days. Performance improvement of such solvers leads to reduced time consumption, enabling more opportunities to fine-tune parameters.

\begin{figure}[h!]
    \begin{center}
    \begin{subfigure}[h]{0.45\textwidth}
        \begin{subfigure}[h]{0.48\textwidth}
            \includegraphics[width=\textwidth]{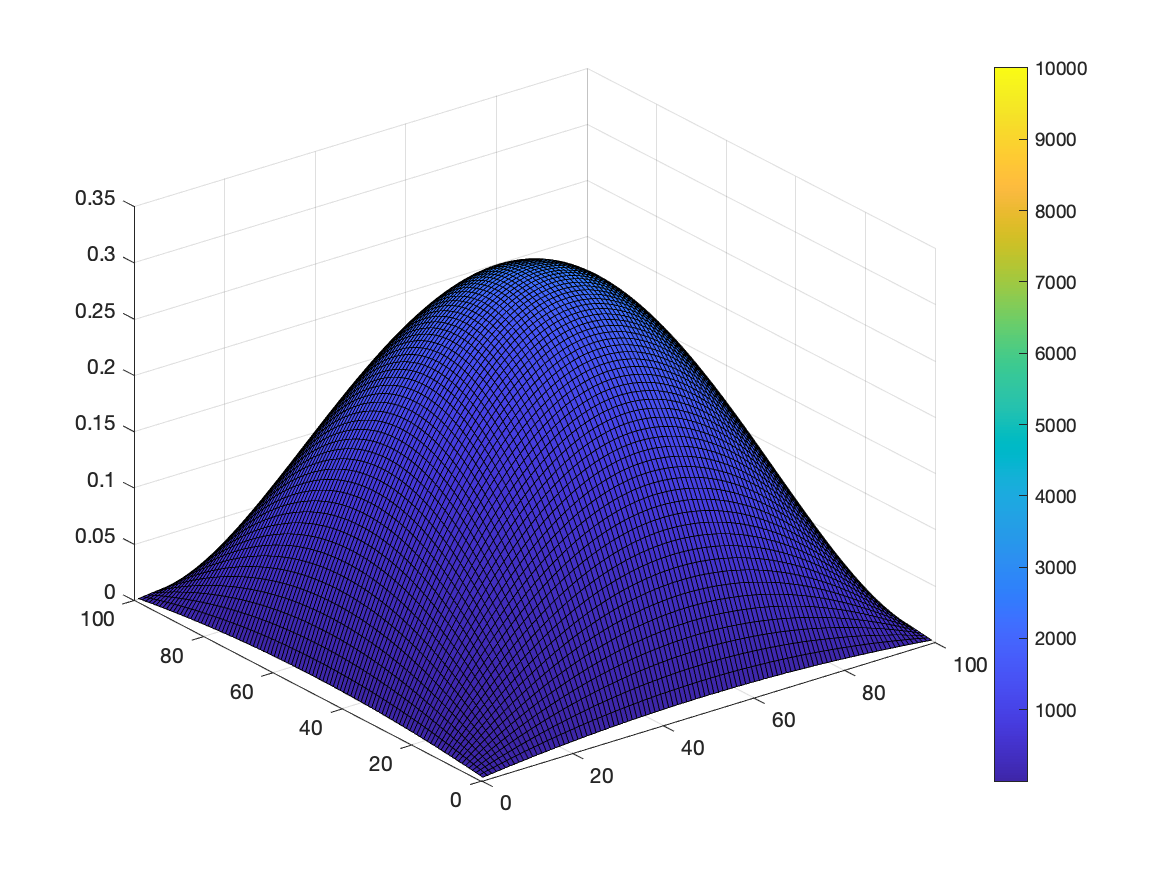}
            \caption*{Desired}
        \end{subfigure}
        \begin{subfigure}[h]{0.48\textwidth}
            \includegraphics[width=\textwidth]{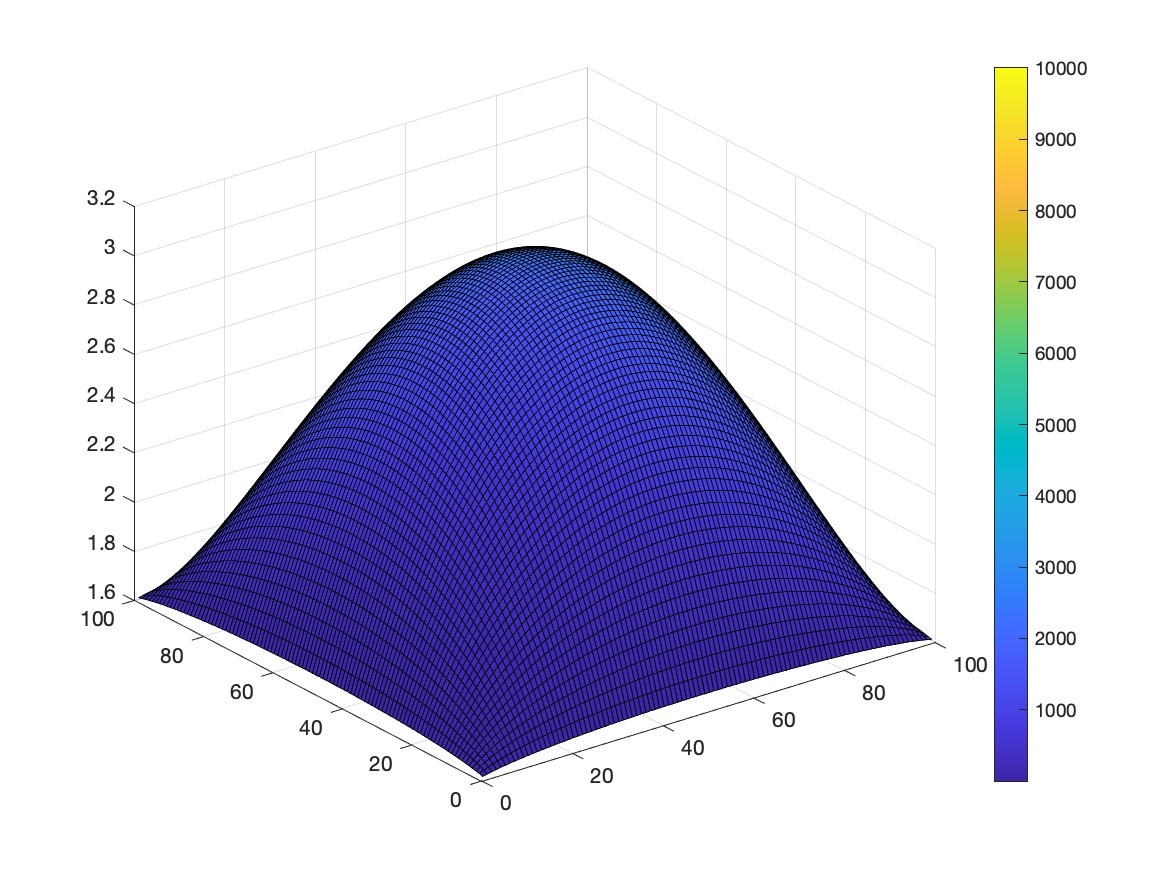}
            \caption*{Optimal solution}
        \end{subfigure}
        \caption{Simple monotonous function}
        \label{fig:simple}
    \end{subfigure}
    \begin{subfigure}[h]{0.45\textwidth}
        \begin{subfigure}[h]{0.48\textwidth}
            \includegraphics[width=\textwidth]{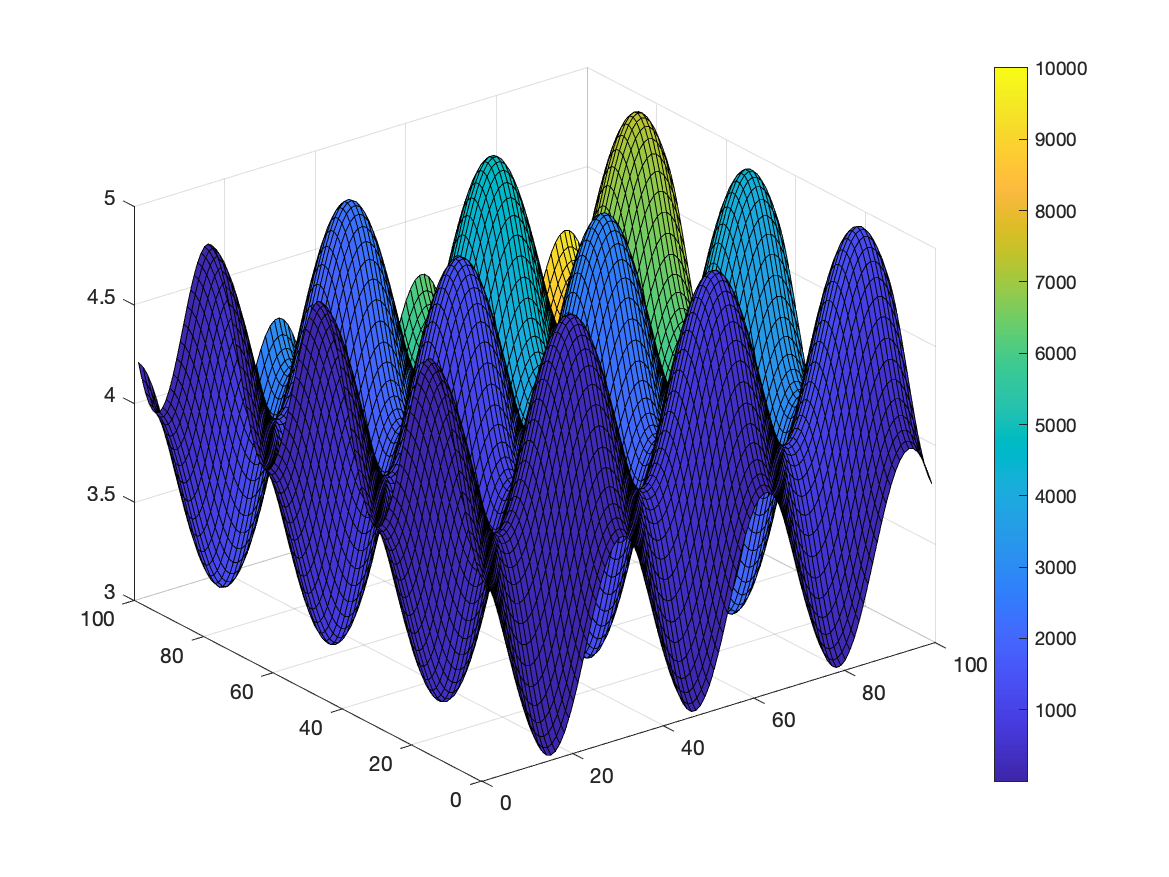}
            \caption*{Desired}
        \end{subfigure}
        \begin{subfigure}[h]{0.48\textwidth}
            \includegraphics[width=\textwidth]{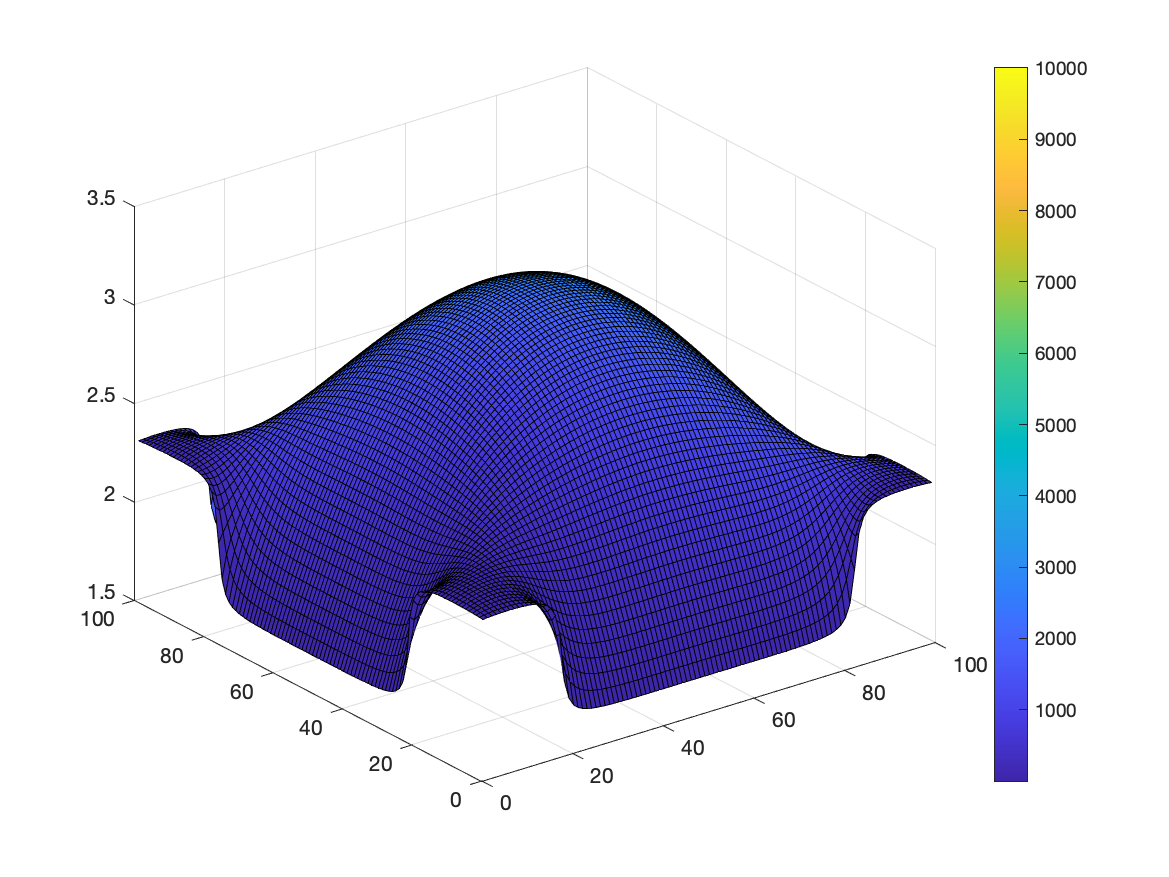}
            \caption*{Optimal solution}
        \end{subfigure}
        \caption{Highly nonlinear function}
        \label{fig:complex}
    \end{subfigure}
    \caption{Governing function of the desired profile and optimal solution obtainable}
    \label{fig:simpleAndComplex}
    \end{center}
\end{figure}

Boundary control problems involve controlling the values at the boundaries while adhering to control and state constraints. The objective is to find the optimal boundary values such that the values in the domain are as close as possible to their desired values, with the closeness defined using residual sum of squares. For simple desired profiles, matching the domain values may be attainable, as in Figure \ref{fig:simple}, while for complicated ones, the best achievable could be far from the desired, as in Figure \ref{fig:complex}.

Many classical approaches and tools use Interior-point method (IPM), a non-linear optimization method, to solve such an optimization problem. IPMs typically require factorization and the solution of a linear system involving the Hessian matrix in each iteration. For large-scale problems, computation of the factorization can be computationally expensive and requires excessive storage, thus dominating the overall complexity of the algorithm. Additionally, as non-convex optimization problems can have multiple local minima, IPMs are not guaranteed to find a global minimum as these are sensitive to the starting point. These limitations serves as the motivation for exploring methodologies in deep learning and reinforcement learning to solve such problems.

\section{Literature review}

The paper by \citesubject{maurer2000optimization} presents control problems for semi-linear elliptic equations subject to control and state constraints. They approached the problem by transforming the control problem into a non-linear programming problem and used the interior-point method to solve the same. An analytical setting for optimal boundary control of one-dimensional heat equations is described by \citesubject{lang2023exact}. Non-elliptic boundary control problems using the Cahn-Hilliard equation as governing PDE exist in literature \cite{colli2015boundary}.

A plethora of studies on using deep learning and reinforcement learning to expedite numerical computing tasks exist. Scientific simulations involving solving partial differential equations (PDEs) can be solved using neural networks \cite{lagaris1998artificial, jiang2023neural, brandstetter2022message, chamberlain2021beltrami, 9302647}. \citesubject{AlphaTensor2022} used reinforcement learning to arrive at a matrix multiplication method that improves upon Strassen's algorithm, which was considered the most optimized for the past fifty years. Deep reinforcement techniques are used on optimization and control problems \cite{williams1992simple, zhu2018optimal}. Attempts to analyze and improve optimizer methods following gradient descent are researched \cite{DBLP:journals/corr/abs-1709-07417, liu2023sophia}. A few papers describing the use of graph neural networks in control are of interest \cite{chen2021graph, shen9024538, meirom2021controlling, rozemberczki2021pytorch, pmlr-v144-gama21a}. Observe a combination of Q-learning with policy-gradients with soft actor-critic method in the work by \citesubject{pmlr-v80-haarnoja18b}. These methods are combined to solve real-world problems \cite{BONNY2022115443}.

Control and optimization studies without deep learning or reinforcement learning also exist. \citesubject{zafar2023structured} discusses a preconditioned conjugate gradient method for solving finite-horizon linear-quadratic optimal control problems. Several studies and tools using the interior-point method can be observed in the literature \cite{tasseff2019exploring, 8846109, KARDOS2022108613, pacaud2023parallel, wachter2006implementation}.

\section{Methodology}

\subsection{High-level architecture}

\begin{figure}[h!]
    \begin{center}
    \includegraphics[width=0.45\textwidth]{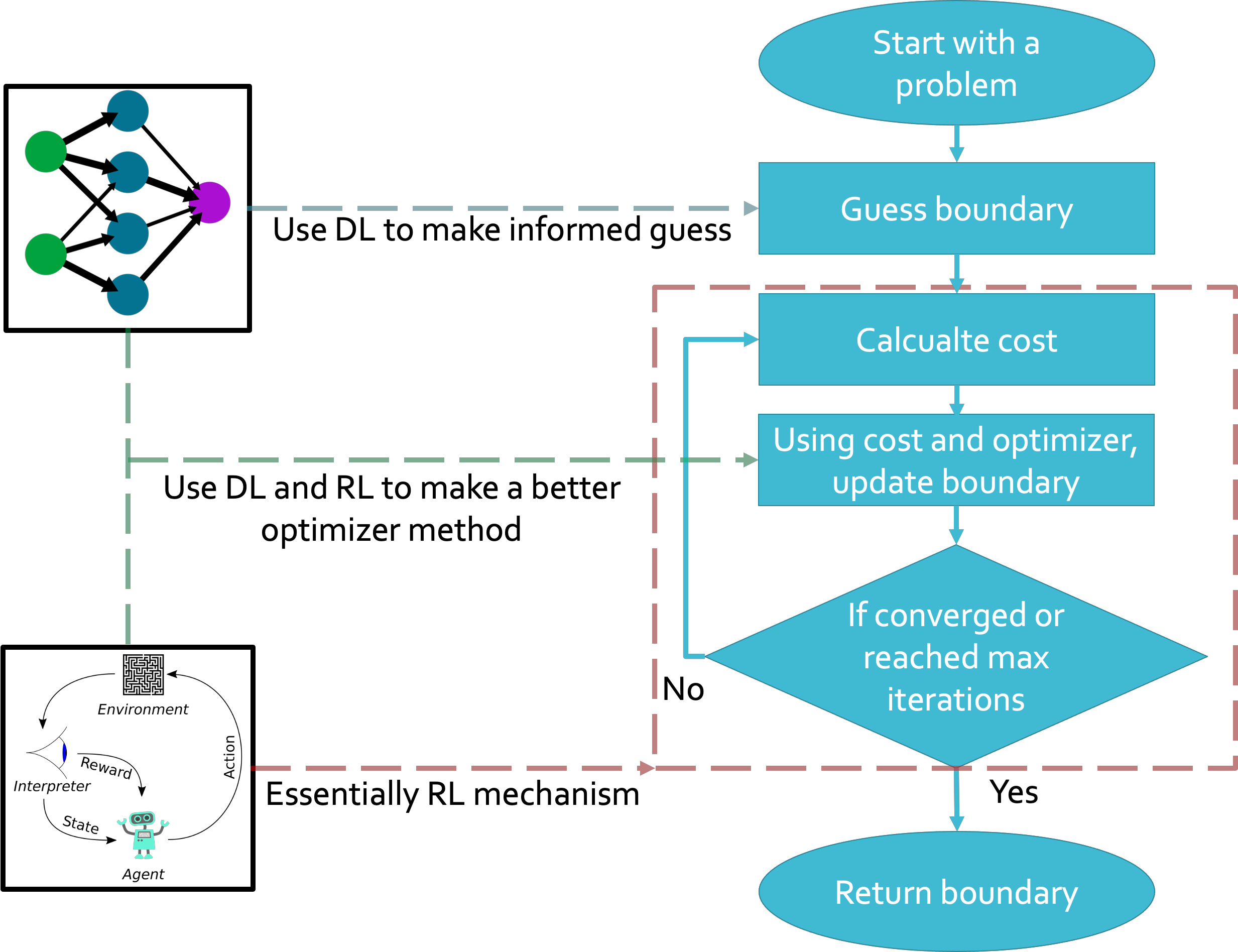}
    \caption{High-level solution architecture}
    \label{fig:highLevelArchitecture}
    \end{center}
\end{figure}

Copious iterative optimization algorithms have a common feature of starting with an initial guess and iteratively updating the current point until reaching the best-fitting solution. As depicted in Figure \ref{fig:highLevelArchitecture}, the proposed method first makes an informed initial guess after considering problem parameters using deep learning, and subsequently, it iteratively updates the solution using an optimizer created using policy gradient reinforcement learning.
From this point forward, the part of the proposed method that makes an informed intial guess is referred to as \textit{initial guess method} and that which performs iterative update is called \textit{optimizer method}.

\subsection{Defining cost and reward}

Mathematically, the boundary control problem is defined as:
\small
\begin{equation*}
    \begin{aligned}
        \underset{y, u}{\textbf{Minimize}} \ &
        \begin{aligned}
        & \frac{1}{2} \int\limits_{\Omega} \left( y(x) - y_d(x) \right)^2 dx + \frac{\alpha}{2} \int\limits_{\Gamma} \left( u(x) - u_d(x) \right)^2 dx \\
        \end{aligned} \\
        \textbf{Subject to} \ & \nabla^2 y = c, \ \ y_{min} < y < y_{max}, \  \ u_{min} < u < u_{max}
    \end{aligned}
\end{equation*}
\normalsize
where $\Omega$ is the domain, $\Gamma$ is the boundaries, $y_d$ is the desired values for the domain, $u_d$ is the desired values for the boundaries, $y$ is the values for the domain, $u$ is the values for the boundaries, $\alpha$ is a non-negative constant that determines how much weight must be given to the cost for the boundaries, $c$ is a constant sourcing term, $y_{min}$ and $y_{max}$ are upper and lower bound of $y$, and, $u_{min}$ and $u_{max}$ are upper and lower bound of $u$.

There are constraints associated with the objective function. To deal with the governing PDE, structure the proposed method to output the boundary values and calculate the domain values using a numerical PDE solver. Along with the objective function, the bound constraints are incorporated into the cost function using a barrier function. The cost function is mathematically defined as:
\small
\begin{align*}
    F &= F_{o} + \beta F_{v} \\
    F_{o} &= \frac{1}{2} \int\limits_{\Omega} \left( y(x) - y_d(x) \right)^2 dx + \frac{\alpha}{2} \int\limits_{\Gamma} \left( u(x) - u_d(x) \right)^2 dx\\
    F_{v} &= \int\limits_{\Omega} f_{\Omega}(x) dx + \int\limits_{\Gamma} f_{\Gamma}(x) dx \\
    f_{\Omega}(x) &=
        \begin{cases}
            (y(x) - y_{min})^2 & \text{if } y(x) \in (-\infty, y_{min}) \\
            (y(x) - y_{max})^2 & \text{if } y(x) \in (y_{max}, \infty) \\
            0 & \text{otherwise}\\
        \end{cases}
        \\
    f_{\Gamma}(x) &=
        \begin{cases}
            (u(x) - u_{min})^2 & \text{if } u(x) \in (-\infty, u_{min}) \\
            (u(x) - u_{max})^2 & \text{if } u(x) \in (u_{max}, \infty) \\
            0 & \text{otherwise}\\
        \end{cases}
\end{align*}
\normalsize
where $\beta$ is the penalty factor for constraint violation, which should have a large positive value.

The reward could be any strictly monotonous decreasing function of the cost function, as maximizing it leads to minimizing the cost. For simplicity, the reward function shall be the negative of the cost function.

\subsection{Data generation}

We scope our study to boundary control problems with Dirichlet boundary conditions. Due to the unavailability of a ready-to-use dataset containing real-world information, an equation generator is created based on four boundary control problems with Dirichlet boundary condi- tions presented in the literature \cite{maurer2000optimization}. To randomly create reasonable problems, it must be ensured that the parameters and coefficients can only vary within a well-defined range, as detailed below.

\begin{itemize}
    \item \textbf{Domain size}: A random integer between $10$ and $100$.
    \item \textbf{Alpha factor}: Set to $0.01$ for all problems.
    \item \textbf{Target profile equation}: The desired values in the domain are calculated using this equation. It shall have quadratic and sine-squared terms in both directions; the coefficients for each of the quadratic terms are integer values between $-5$ and $5$ and the frequency and the phase angles of the sine squared terms are respectively set to be an integral multiple of $\pi$ between $-5$ and $5$, and, an integral fraction of $\pi$ between $1$ and $6$. Examples:
    \begin{itemize}
        \item $x_2^2 - x_2 + \sin^2(2\pi x_1 + \frac{\pi}{5})$
        \item $\sin^2(3\pi x_1 + \frac{\pi}{6}) + \sin^2(\pi x_2 + \pi)$
        \item $2x_2^2 + x_2$
    \end{itemize}
    \item \textbf{Desired boundary values}: Set to zero for all problems.

    \item \textbf{Bounds}: The domain lower bound in all of the problems in the original set of problems was $-10^{20}$, which signifies negative infinity, which is used in the problems generated here as well. The rest of the bound values are chosen based on the target profile equation; by generating the desired domain profile by solving the target profile equation for the domain size. The maximum, minimum and median values in the desired domain are used.

    The following are the specifications of the Bounds
    \begin{itemize}
        \item \textit{Lower bound for domain} is set to $-10^{20}$ for all problems.
        \item \textit{Upper bound for domain} is a uniformly sampled random number between the median and the maximum values.
        \item \textit{Lower bound for boundary} is the minimum value plus half of uniform random value generated between positive and negative difference between maximum and minimum values.
        \item \textit{Upper bound for boundary} is the maximum value plus half of uniform random value generated between positive and negative difference between maximum and minimum values.
    \end{itemize}

    \item \textbf{Sourcing term}: A random value chosen from the set $\{ 0, -10, -20, -30, -40, -50 \}$.
    \item \textbf{Additional filters}: Once the problem is generated, there are two additional sets of filters based on certain thresholds that were incorporated to improve the quality of the equations generated. The first one is a threshold for the maximum and minimum values observed in the domain; a difference of less than $0.3$ is discarded. The second one is a threshold on the cost that IPOPT predicts for the generated problem; a problem with a cost of more than $0.2$ per cell is discarded.
\end{itemize}

\section{Experimental setup}

\subsection{Setting up the baselines}

Two separate sets of baselines are defined to assess the effectiveness of the initial guess and optimizer methods.

\subsubsection{Baselines for initial guess method}

Three baselines for initial guess are created with the boundaries set to the mean, the median or the values at the edges of the desired domain, and calculating the cost. This is done for all the generated problems, and the information is stored to make it easier for analysis later.

\subsubsection{Baselines for optimizer method}

Two baselines are used to evaluate the optimizer method.
\begin{enumerate}
    \item \textit{Comparison against solvers previously tried and tested to work for boundary control problems}. The state-of-the-art large-scale nonlinear optimization problem solver, IPOPT \cite{wachter2006implementation}, is used to create this baseline.

    \item \textit{Comparison against optimizers used in deep learning}. On a network with only a single bias layer with the size of the boundaries, backpropagation is performed using the gradients computed from cost calculation. After several iterations, the bias layer is expected to achieve an optimal boundary value. Two different baselines are generated using this approach with Stochastic gradient descent (SGD) and adaptive moment estimation (Adam) as optimizers, and running for $100$ optimizer steps.
\end{enumerate}

Note that the former baseline method is computationally less expensive and can run for all the generated problems. However, since the latter is computationally expensive, the baseline is generated with only $400$ problems.

\subsection{Performing the experiments}

For both initial guess and optimizer methods, neural networks are designed based on intuition, which may involve some feature engineering. The dataset is divided in the ratio 80:10:10 for training, validation and testing. The best model for a given network design is chosen based on the lowest validation cost, while the overall best network design is selected based on the lowest testing cost. The idea is to iteratively design, train, validate and test different networks until models that outperforms the baselines are found.

\subsubsection{Initial guess method}

Inputs to the initial guess method are the desired domain profile, bound constraints and the sourcing term, and the output is the guessed boundary values. While designing the network, ensure that the input and output dimensions are variable and the output dimension is a function of input dimensions. If the target profile array is of size $N \times N$, then the output boundary value size would be $4 \times N$. Intuitively, one could use convolution, graph convolution and different pooling layers in the network. Indeed, there may be other layers that could work with this constraint. A fully connected layer, a.k.a. linear layer, can not work with such a constraint as it requires the input and output dimensions to be known beforehand to work.

\subsubsection{Optimizer method}

Inputs to the optimizer method are the boundary values and the calculated gradients with respect to the cost, and the output is updated boundary values. All of these have size $4 \times N$. Inspired by several widely used optimizers, the network may have the ability to work with both spatial and temporal information. The temporal features would help in taking the previous iteration into consideration. Hence, layers like convolution, graph convolution, recurrent and other layers that can work with spatial and temporal information shall be employed.

\section{Results}

\subsection{Initial guess method}

\subsubsection{Architecture}

\begin{figure}[h!]
    \begin{center}
    \includegraphics[width=0.45\textwidth]{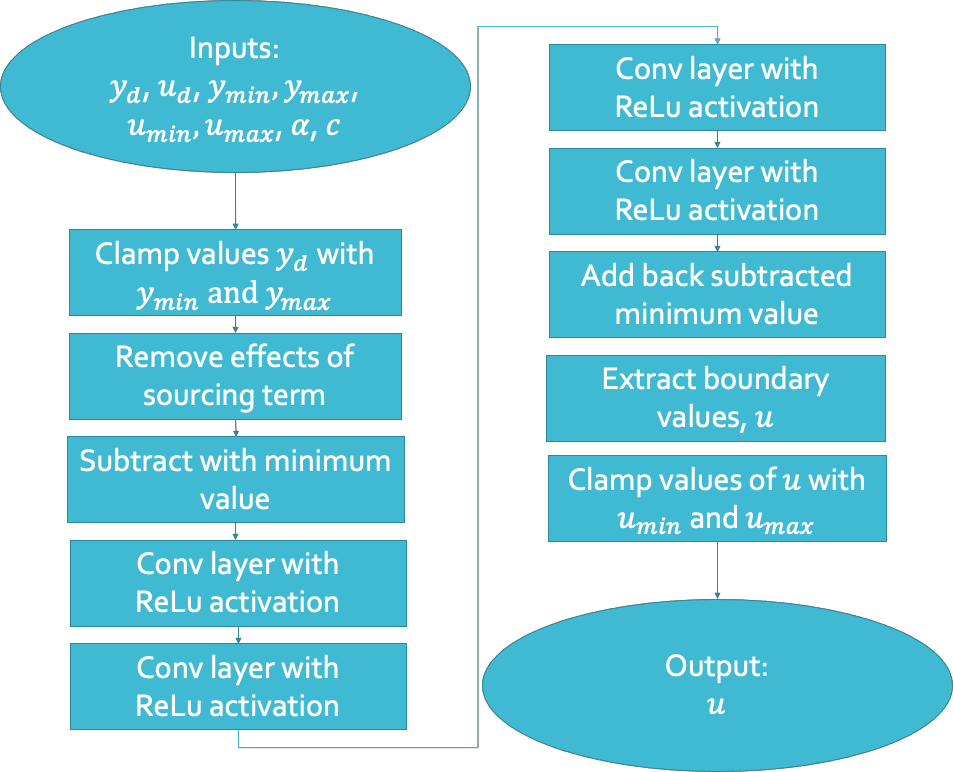}
    \caption{Network architecture for initial guess method}
    \label{fig:initialGuessNetworkArchitecture}
    \end{center}
\end{figure}

After designing and evaluating several networks, the architecture of the best one obtained is shown in Figure \ref{fig:initialGuessNetworkArchitecture}.
Starting with the desired domain profile (a 2-dimensional array), the values are clamped to the domain's upper and lower bounds. From this, the sourcing term effect is subtracted. Following this, to ensure that all the values are greater than or equal to zero, the minimum value in the array is determined, and all the elements are subtracted with this minimum value. The resulting array passes through four convolution layers with rectified linear unit activations, following which the previously subtracted minimum value is added back. The boundaries of the resulting array are extracted and clamped to the boundary's upper and lower bounds. The result forms the guessed boundary values.

\subsubsection{Quantitative}

Table \ref{tab:ResultsSummary} summarizes the comparison between the baselines and the initial guess method filtered for the problems for which IPOPT found feasible solutions.

\begin{figure}[h!]
    \begin{center}
        \begin{subfigure}[h]{0.22\textwidth}
            \includegraphics[width=\textwidth]{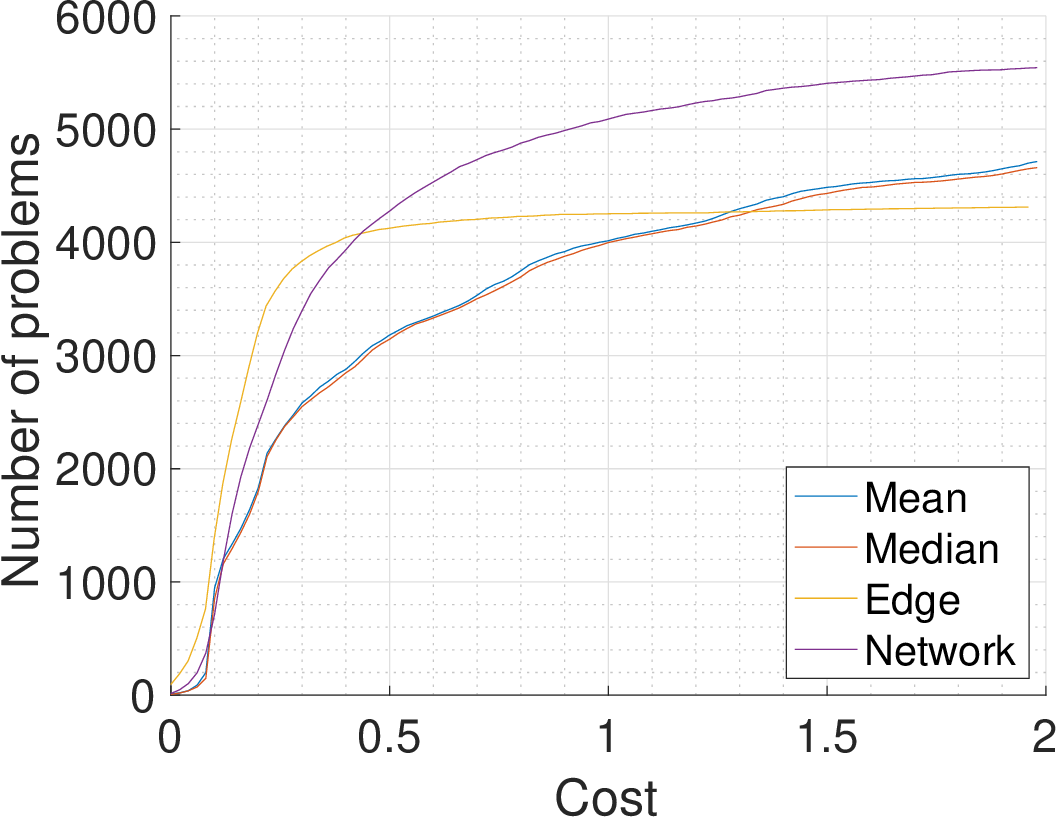}
            \caption{Trimmed till cost value of 2}
        \end{subfigure}
        \begin{subfigure}[h]{0.22\textwidth}
            \includegraphics[width=\textwidth]{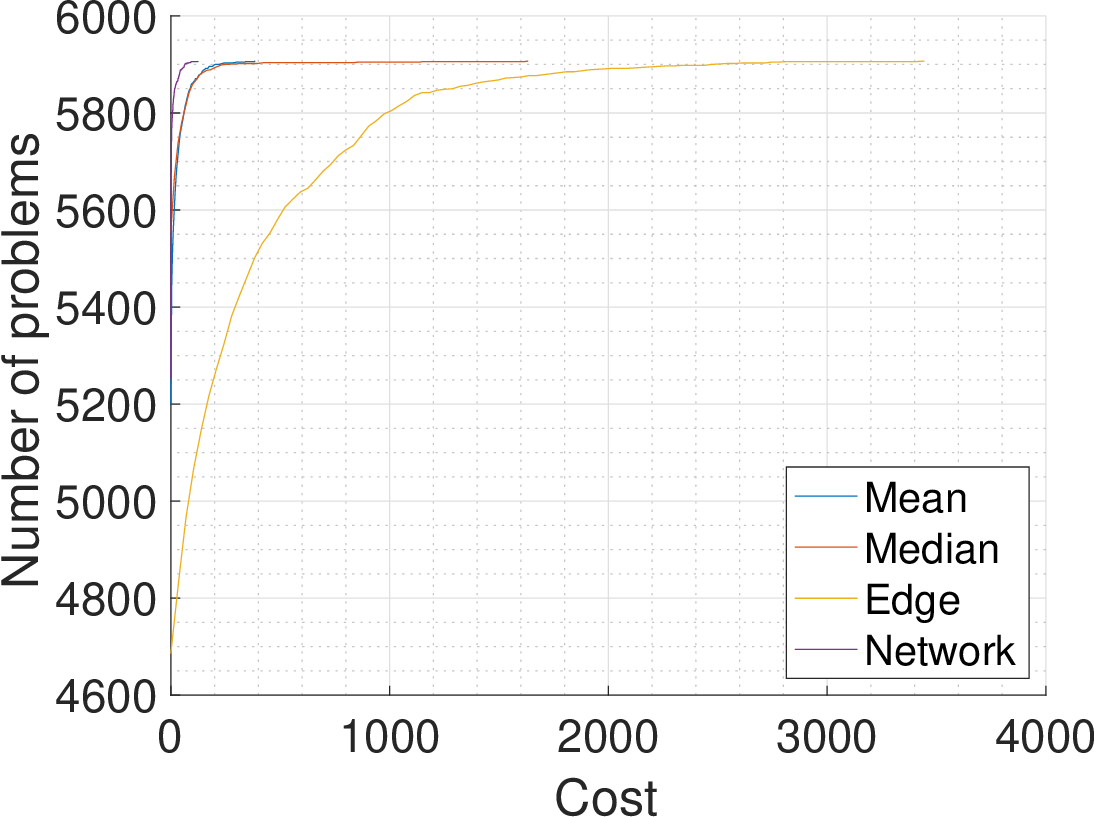}
            \caption{Larger cost values}
        \end{subfigure}
        \caption{Cumulative plot comparing cost for different methods for making initial guess}
        \label{fig:initialGuessCumulativeHistogram}
    \end{center}
\end{figure}

A cumulative histogram plot with information from the baselines and the network is shown in Figure \ref{fig:initialGuessCumulativeHistogram}. The plot depicts the count of problems with costs less than or equal to the value in the X-axis at any given point.
Observe that there are about $4000$ problems for which using edge values may be more advantageous than the other three. However, after the cost value mark of about $0.42$, the method appears to have the upper hand. From the sub-plot for larger cost values, it is evident that the edge method performs significantly worse than the other three methods at later points as it goes slowly into the regions with more and more cost.

Although it is evident that the proposed initial guess method performs better than baselines by mean and median values of the desired profile, it is inconclusive as to whether it is better than that of the edge values. However, it can be argued that using the edge values as a starting point is unreliable as it could be the best or the worst in different situations; therefore, the use of the more consistent initial guess method is advisable.

\begin{table*}
    \centering
    \begin{tabular}{|c||c|c|c|c||c|c|c|c|c|}
        \hline
        & \multicolumn{4}{c||}{\textbf{Initial guess}} & \multicolumn{5}{c|}{\textbf{Iterative optimization}}\\
        \hline
        & \textbf{Mean} & \textbf{Median} & \textbf{Edge} & \textbf{Method} & \textbf{IPOPT} & \textbf{SGD} & \textbf{Adam} & \multicolumn{2}{c|}{\textbf{Method}} \\
        \hline
        \textbf{Mean cost} & 4.5059 & 5.6269 & 87.1563 & 1.0591 & 0.1223 & 8378.6 & 0.1808 & 0.1721 & 0.1223 \\
        \textbf{Median cost} & 0.4422 & 0.4572 & 0.2010 & 0.2709 & 0.1252 & 0.3439 & 0.1672 & 0.1469 & 0.1181 \\
        \textbf{Lowest cost} & 0.0087 & 0.0091 & 0.000023215 & 0.0081 & 0.0097 & 0.0093 & 0.0164 & 0.0100 & 0.0034 \\
        \textbf{Highest cost} & 388.7509 & 1642.7 & 3478.2 & 125.8895 & 0.1995 & 126160 & 0.5770 & 1.0894 & 0.6616 \\
        \hline
        \textbf{Iterations} & & & & & $43^*$ & 100 & 100 & 8 & 32 \\
        \hline
        \multicolumn{10}{c}{\quad\footnotesize$^*$ The number of iterations for IPOPT is the rounded mean value}
    \end{tabular}
    \caption{Summary of method costs compared with the baselines}
    \label{tab:ResultsSummary}
\end{table*}

\subsection{Optimizer method}

\subsubsection{Architecture}

\begin{figure}[h!]
    \begin{center}
    \includegraphics[width=0.40\textwidth]{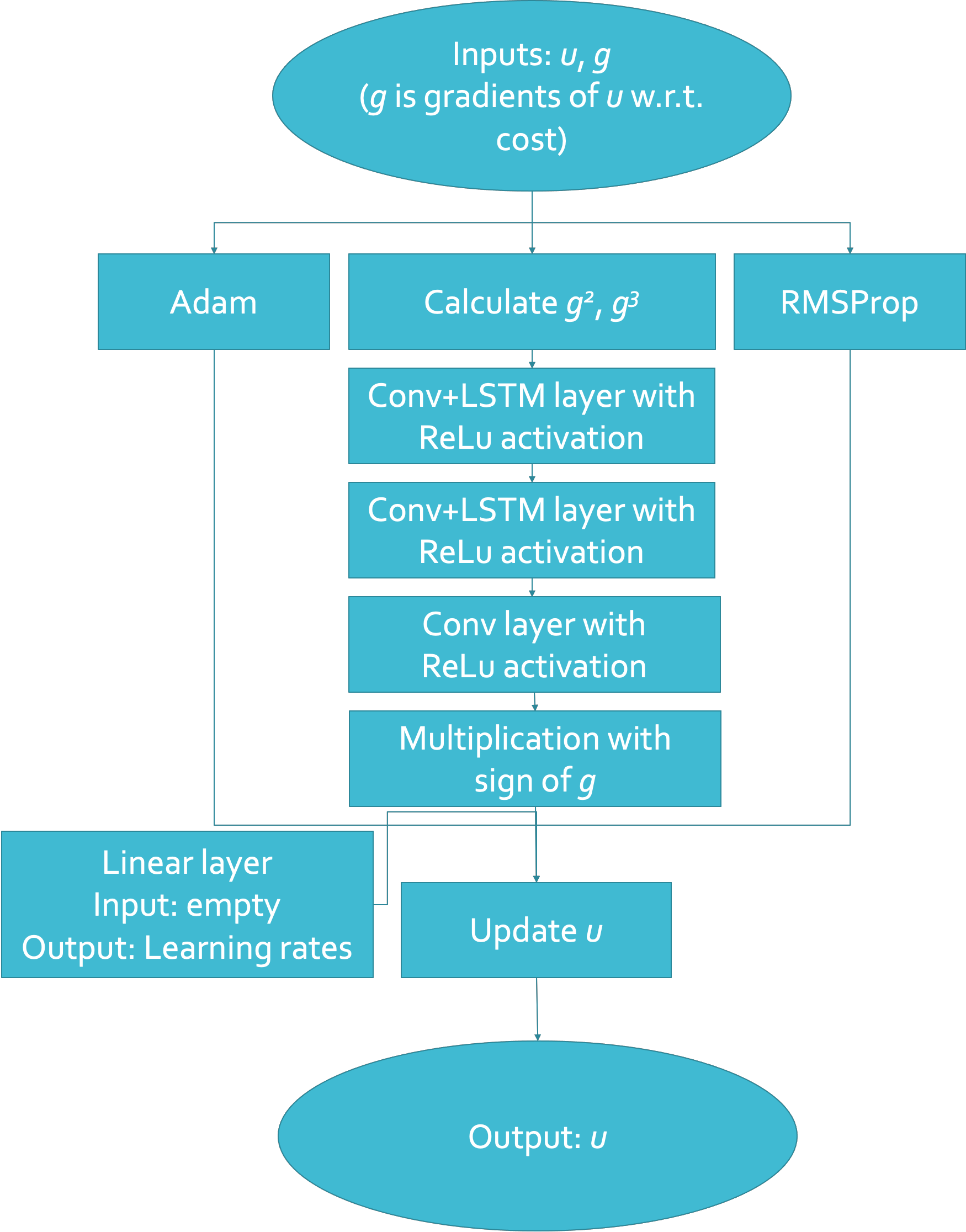}
    \caption{Network architecture for optimizer method}
    \label{fig:optimizerNetworkArchitecture}
    \end{center}
\end{figure}

Figure \ref{fig:optimizerNetworkArchitecture} shows the architecture of the best method obtained after several iterations.
The method encapsulates the Adam optimizer, RMSProp optimizer and a custom spatio-temporal network. The gradient is the input to these three, and the outputs are some intermediary values, which are scaled using three separate learnable learning rates and used to update the boundaries.

The spatio-temporal part begins by concatenating the gradients, and their squared and cubed values, which is passed through a couple of custom temporal convolution layers with rectified linear unit activation and, finally, a convolution layer with rectified linear unit activation. This intermediary output is multiplied with the sign of the original gradient values. Note that the custom temporal convolution layer encapsulates convolution and LSTM layers.

\subsubsection{Quantitative}

Out of the $400$ problems selected, IPOPT found feasible solutions for $252$. The costs corresponding to these problems by IPOPT, SGD and Adam are used as baselines. Comparison between the baselines and the optimizer method run for $8$ and $32$ steps are summarized in Table \ref{tab:ResultsSummary}. During training and validation, the network ran for $8$ steps. The results show that the method achieves lower cost after $32$ iterations, indicating it is indeed iteratively minimising the cost.

\begin{figure}[h!]
    \begin{center}
    \includegraphics[width=0.22\textwidth]{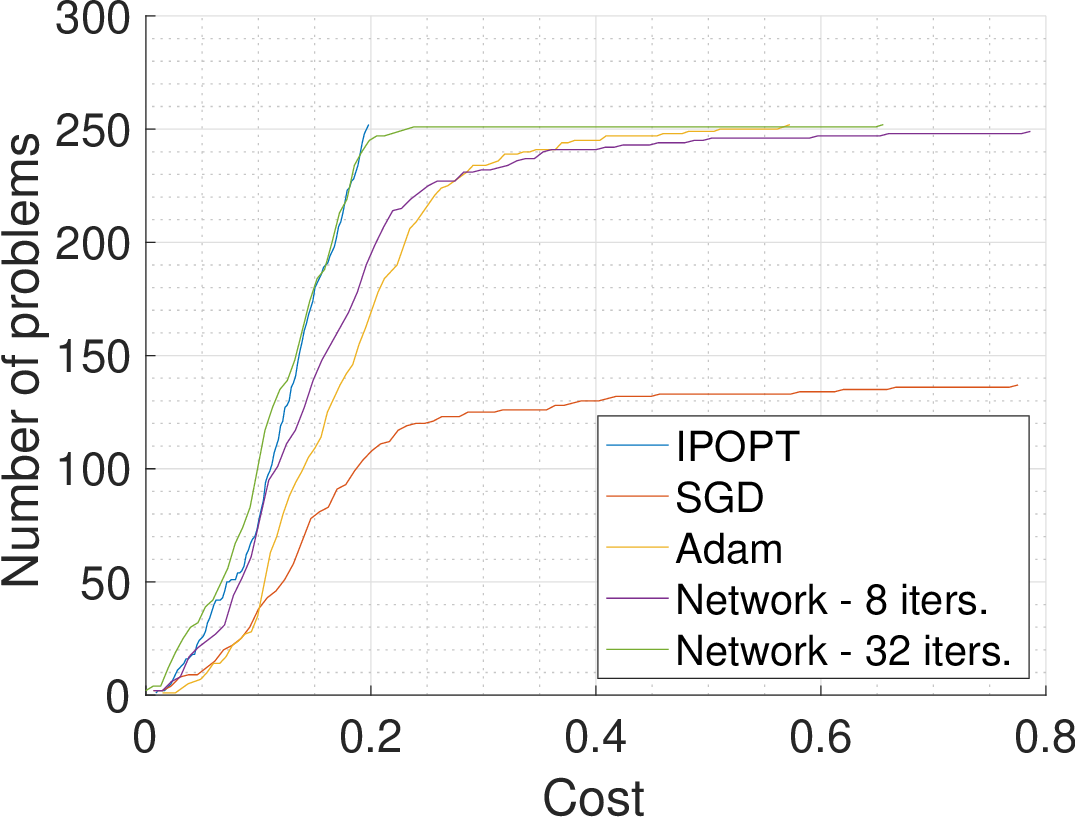}
    \caption{Cumulative plot comparing cost for different optimizers with proposed optimizer method}
    \label{fig:optimizerNetCumulativeHistogram}
    \end{center}
\end{figure}

A cumulative histogram plot comparing the baselines with the optimizer method is shown in Figure \ref{fig:optimizerNetCumulativeHistogram}.
Observe the fiece competition between IPOPT and the optimizer method running $32$ iterations; the lines corresponding to these interset quite a few times.

In a head-to-head comparison between the optimizer method and the baselines, it beats IPOPT, SGD and Adam-based baselines $127$, $226$ and $244$ times, respectively, out of $252$ cases, which means that the network beats the respective baselines $50.40\%$, $89.68\%$ and $96.83\%$ of the times. The method arrived at is indeed quite good.

\section{Analysis}

A detailed analysis is performed using all $10000$ problems and comparing exclusively against IPOPT.

\subsection{Simple statistics}

Out of the $10000$ problems generated, IPOPT found feasible solutions in $5907$ cases. Among these cases, the cost predicted by the proposed method is lower than IPOPT in $3011$ cases, which is $50.97\%$ cases. However, we use a slightly relaxed barrier function for cost calculation. The maximum and mean absolute constraint violation for a particular cell in the domain or the boundary is $0.1033$ and $8.3 \times 10^{-4}$. The total number of cases wherein the method predicted a feasible solution with zero constraint violation is $2127$, which is $36.01\%$ cases.

\subsection{Iteration at which the network finds the best solution}

\begin{figure}[h!]
    \begin{center}
        \begin{subfigure}[h]{0.22\textwidth}
            \includegraphics[width=\textwidth]{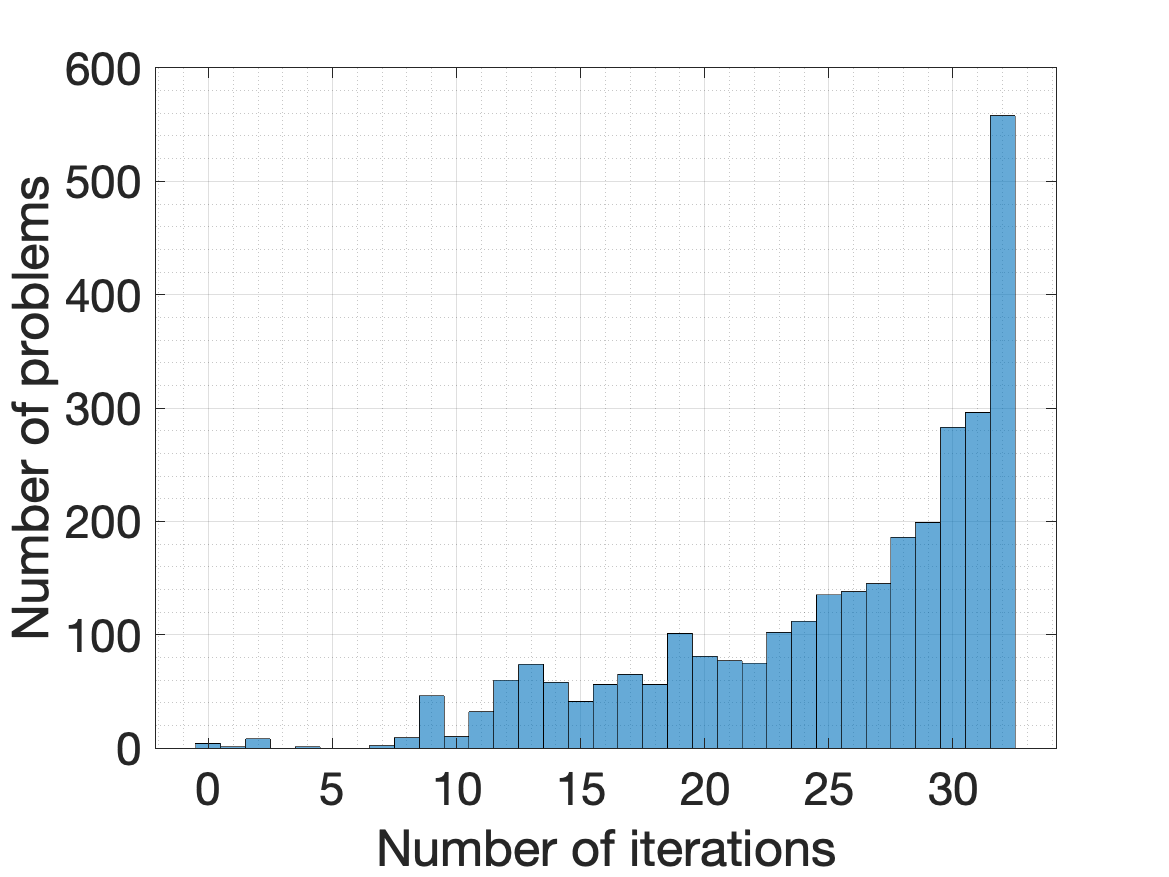}
            \caption{With/without constraint violation}
        \end{subfigure}
        \begin{subfigure}[h]{0.22\textwidth}
            \includegraphics[width=\textwidth]{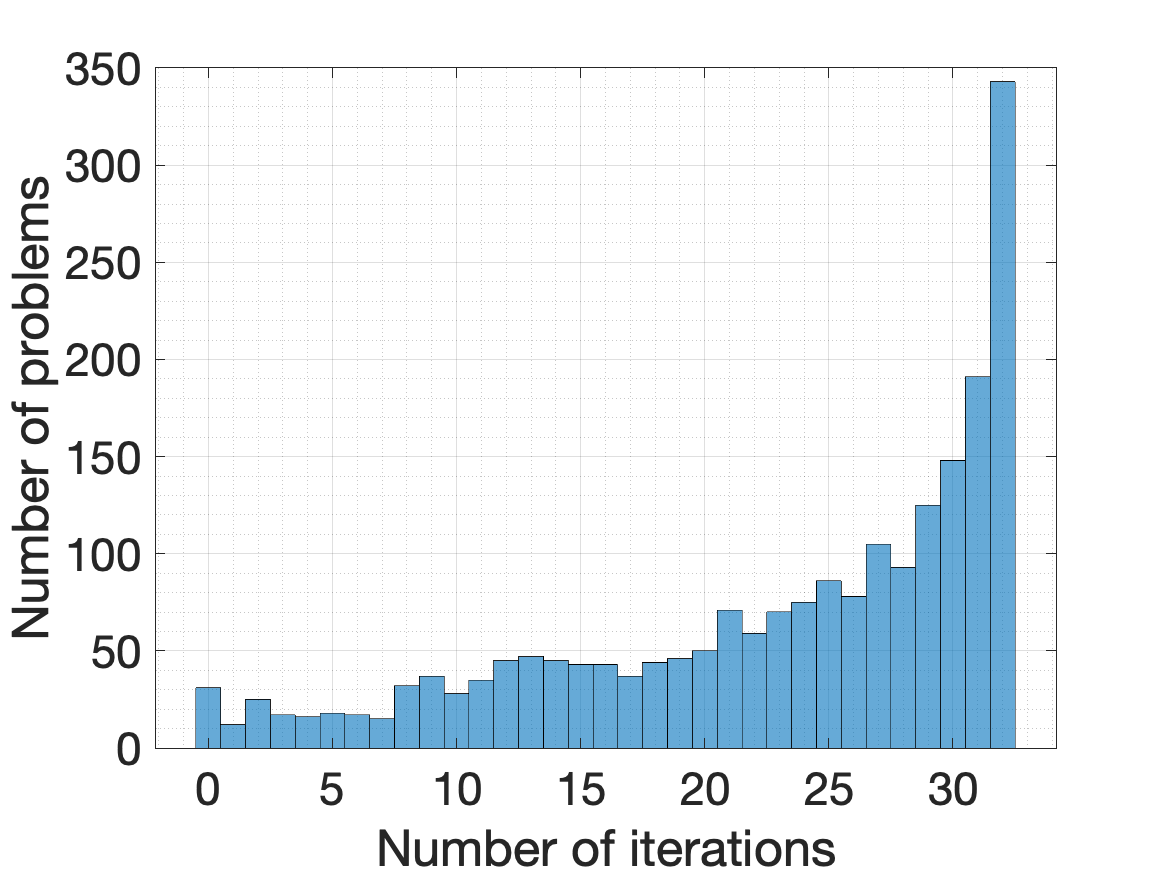}
            \caption{Without constraint violation}
        \end{subfigure}
        \caption{Histograms depicting the number of problems for which the method achieved the lowest cost at which iteration}
        \label{fig:lowestCostIterHistogram}
    \end{center}
\end{figure}

Figure \ref{fig:lowestCostIterHistogram} shows histograms that help visualize the iteration at which the network finds the lowest cost for different problems.
Observe the general trend of obtaining lowest cost at later iterations. The median number of iterations for the lowest cost is $28$ and $26$, respectively, for with and without constraint violation. Furthermore, the tremendous increase in the bar size at iteration number $32$ may suggest that lower costs are obtainable with more iterations for some problems.

\subsection{Iteration at which the network beats IPOPT}

\begin{figure}[h!]
    \begin{center}
        \begin{subfigure}[h]{0.22\textwidth}
            \includegraphics[width=\textwidth]{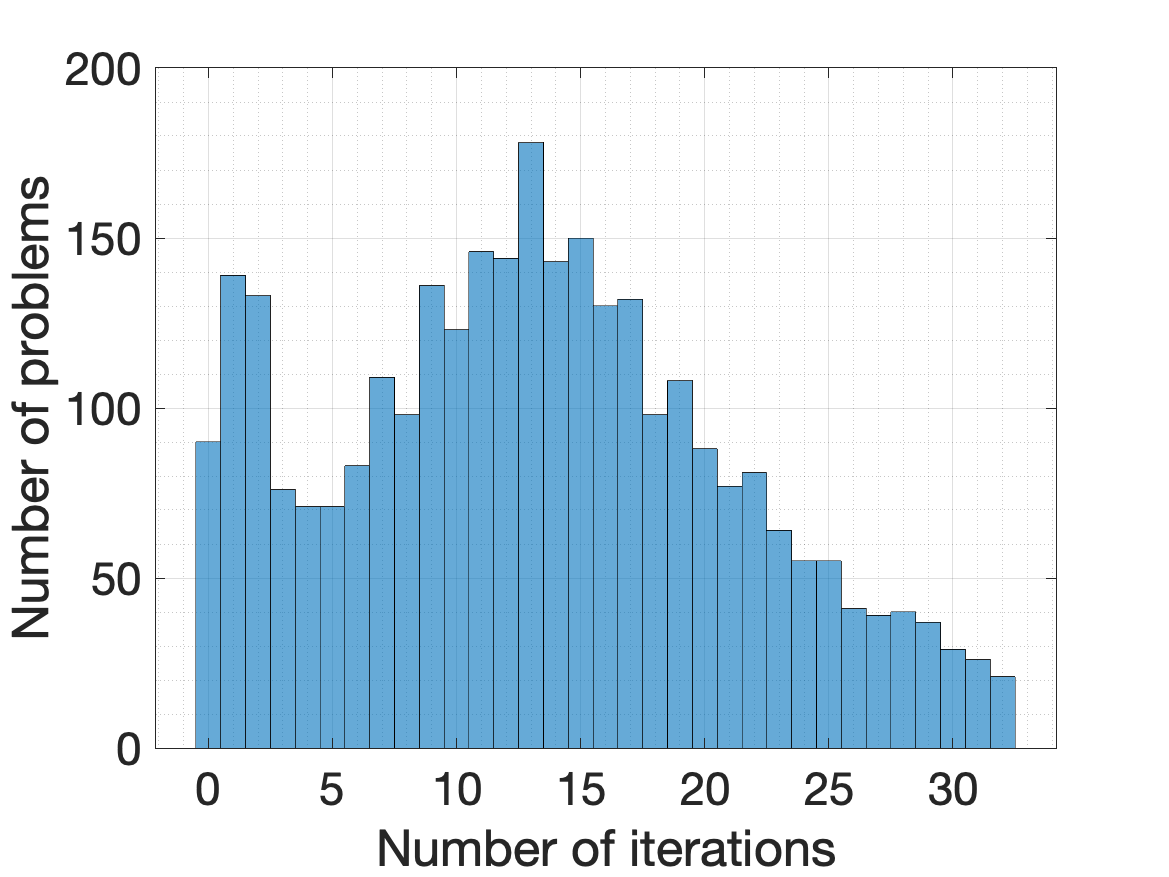}
            \caption{With/without constraint violation}
        \end{subfigure}
        \begin{subfigure}[h]{0.22\textwidth}
            \includegraphics[width=\textwidth]{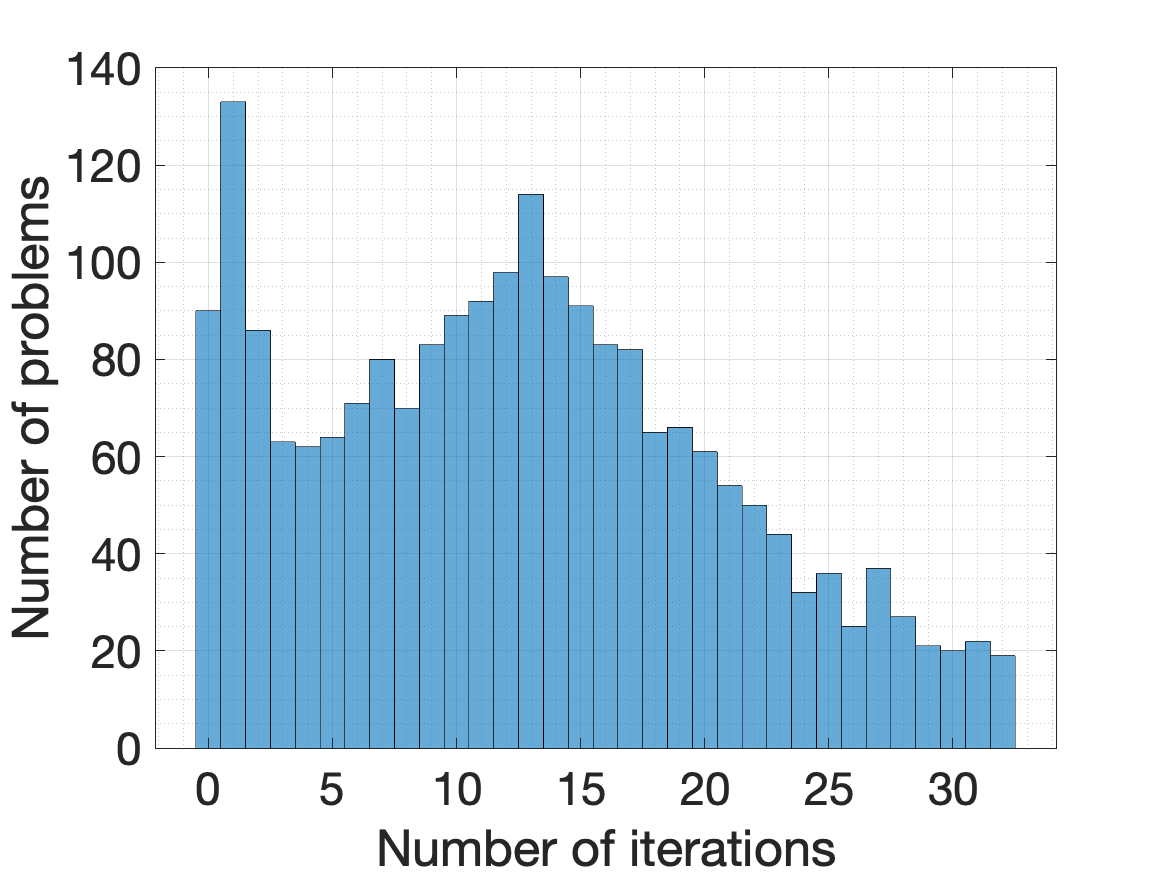}
            \caption{Without constraint violation}
        \end{subfigure}
        \caption{Histograms depicting the number of problems for which the method first surpassed IPOPT's cost at which iteration}
        \label{fig:methodBeatsIPOPTIterHistogram}
    \end{center}
\end{figure}

Histograms that help in visualizing the iteration at which the network achieves lower cost than IPOPT are shown in Figure \ref{fig:methodBeatsIPOPTIterHistogram}.
There are two peaks in both histograms. The former peak attributes the contribution of the initial guess method to the overall results, while the latter shows the prowess of the optimizer method. The median number of iterations to beat IPOPT's cost is $13$ and $12$, respectively, with and without constraint violation. Note that this plot is not informative in cases where the network has a higher cost than IPOPT.

\subsection{Comparing cost and violation with IPOPT}

\begin{figure}[h!]
    \begin{center}
    \includegraphics[width=0.22\textwidth]{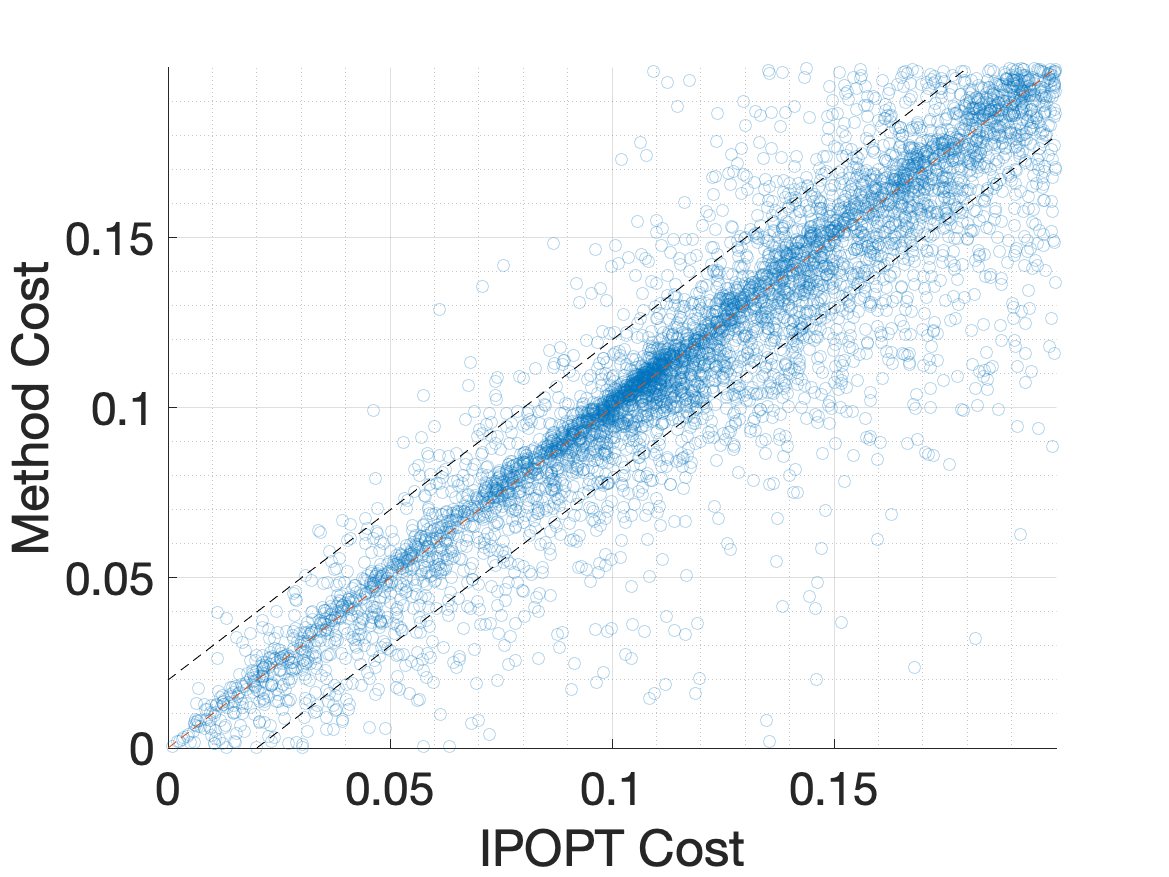}
    \caption{A scatter plot comparing costs}
    \label{fig:OptimizerNetworkVSIPOPT}
    \end{center}
\end{figure}

A semi-transparent scatter plot with the cost by IPOPT in one axis and cost by proposed method in the other is shown in Figure \ref{fig:OptimizerNetworkVSIPOPT}.
The line of equality (in orange) coincides with many scatter plot points, indicating the cost by IPOPT and method to be equal. $4607$ points lie between two parallel lines to the line of equality with X and Y intercepts of $0.02$, respectively, which is nearly $78\%$ of all the points.

\begin{figure}[h!]
    \begin{center}
        \begin{subfigure}[h]{0.22\textwidth}
            \includegraphics[width=\textwidth]{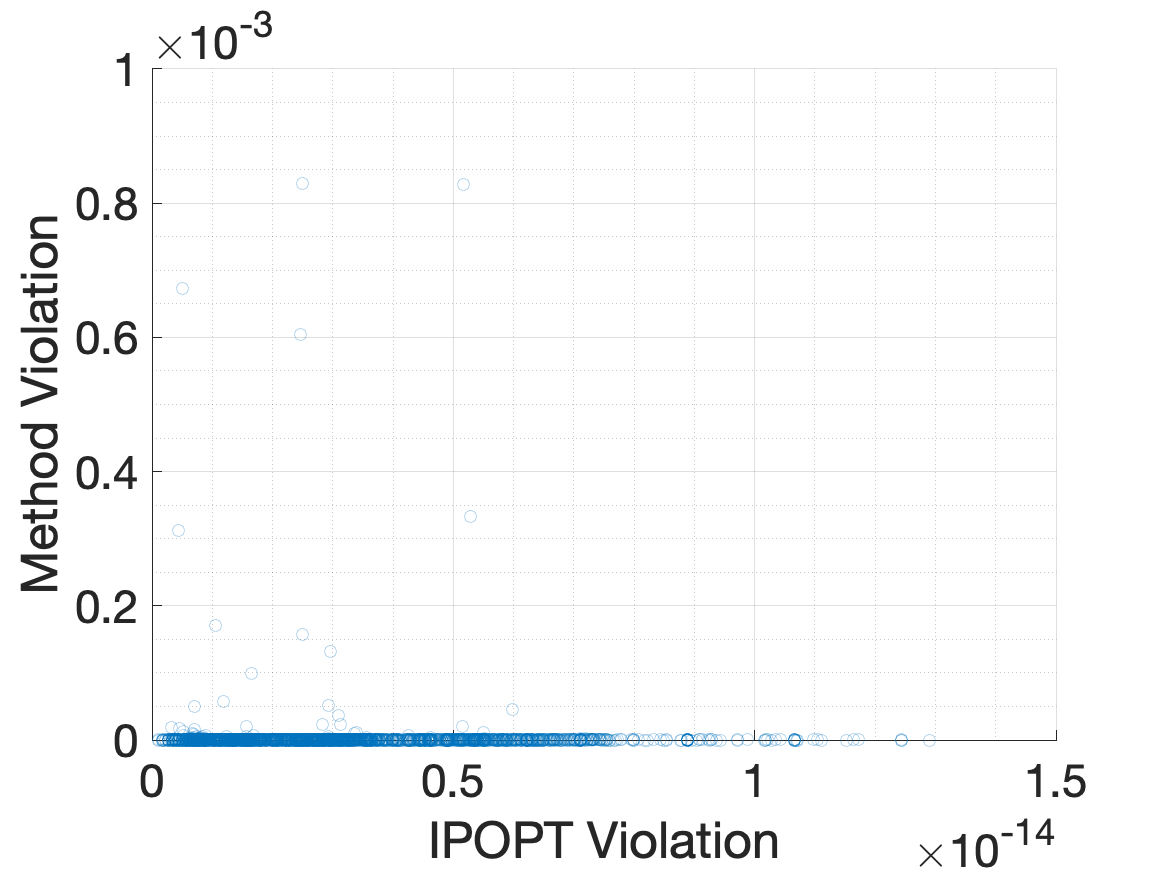}
            \caption{Standard axes}
        \end{subfigure}
        \begin{subfigure}[h]{0.22\textwidth}
            \includegraphics[width=\textwidth]{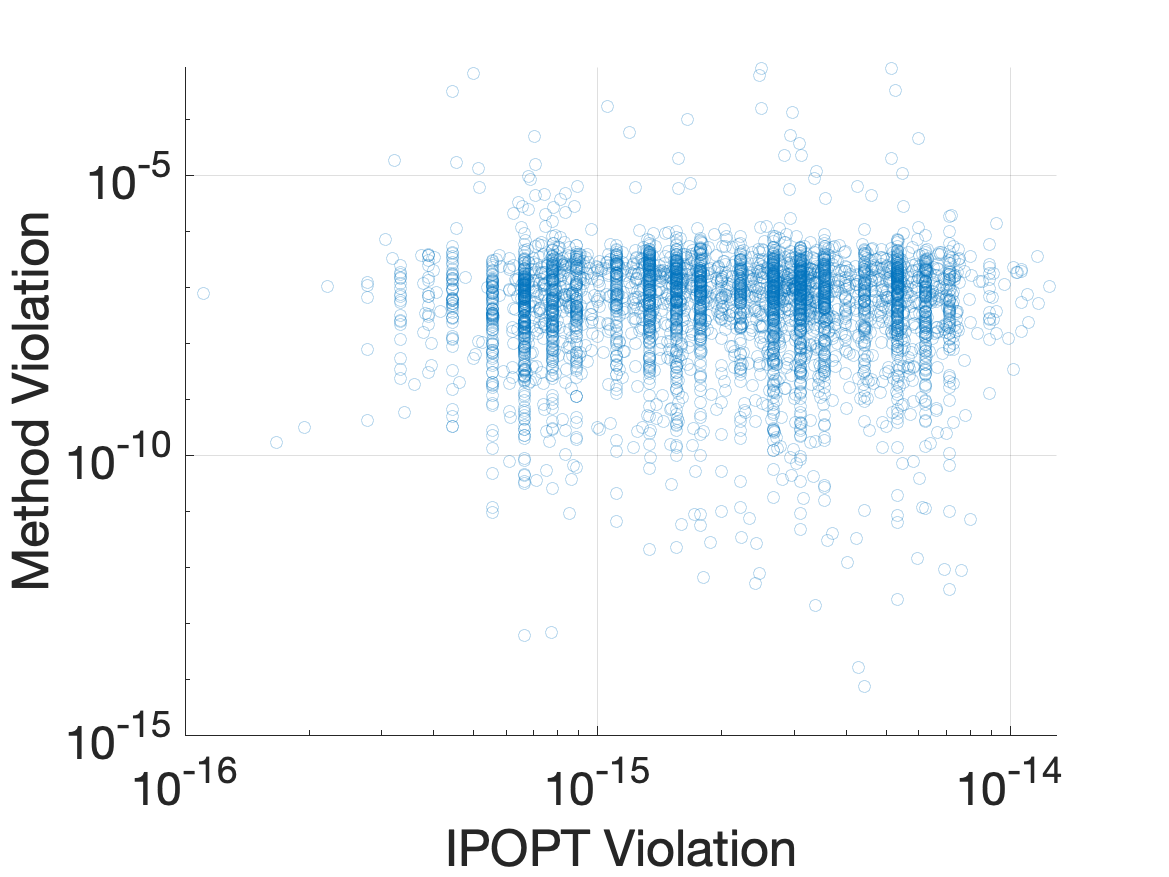}
            \caption{Log-log axes}
        \end{subfigure}
        \caption{Scatter plots comparing constraint violations}
        \label{fig:OptimizerNetworkVSIPOPTViolation}
    \end{center}
\end{figure}

Figure \ref{fig:OptimizerNetworkVSIPOPTViolation} shows a scatter plot for the violations made by IPOPT and the method.
The violation is zero in $1581$ cases for the proposed method, which are not visible in the logarithmic version of the plot. However, there are cases wherein the violations made by the optimizer method are orders of magnitude larger than what is by IPOPT; much of that by IPOPT is in the range between $10^{-15}$ and $10^{-14}$, whereas the same by the optimizer method is in the range between $10^{-8}$ and $10^{-6}$.

\subsection{Performance comparison with IPOPT}

Comparing the performance is a bit tricky for two reasons.
\begin{enumerate}
    \item There are cases where the method achieves lower cost at very early iterations, and there are cases where it does not accomplish the same even after the maximum number of iterations.
    \item The proposed method and IPOPT are implemented in Python and C++, respectively. C++ is between $10$ to $100$ times faster than Python. 
\end{enumerate}
Therefore, take the following analysis with a pinch of salt.

\subsubsection{Timing}

\begin{figure}[h!]
    \begin{center}
        \begin{subfigure}[h]{0.22\textwidth}
            \includegraphics[width=\textwidth]{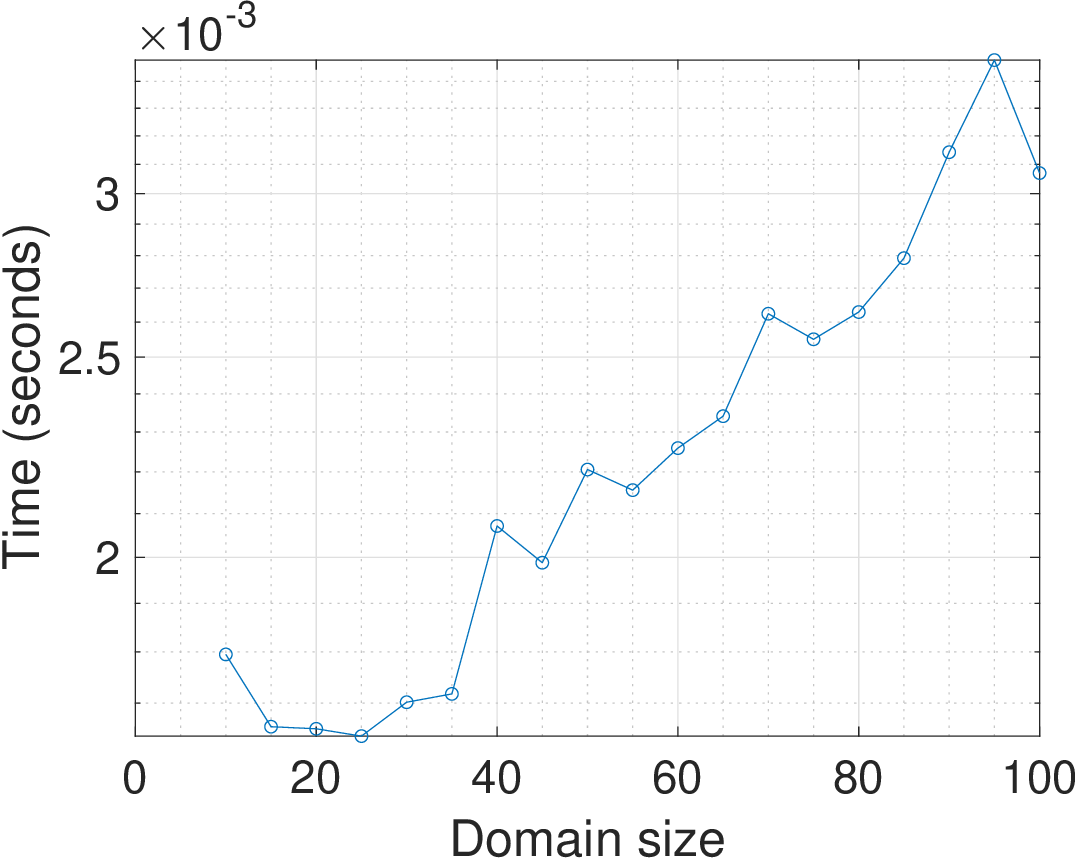}
            \caption{For initial guess method}
        \end{subfigure}
        \begin{subfigure}[h]{0.22\textwidth}
            \includegraphics[width=\textwidth]{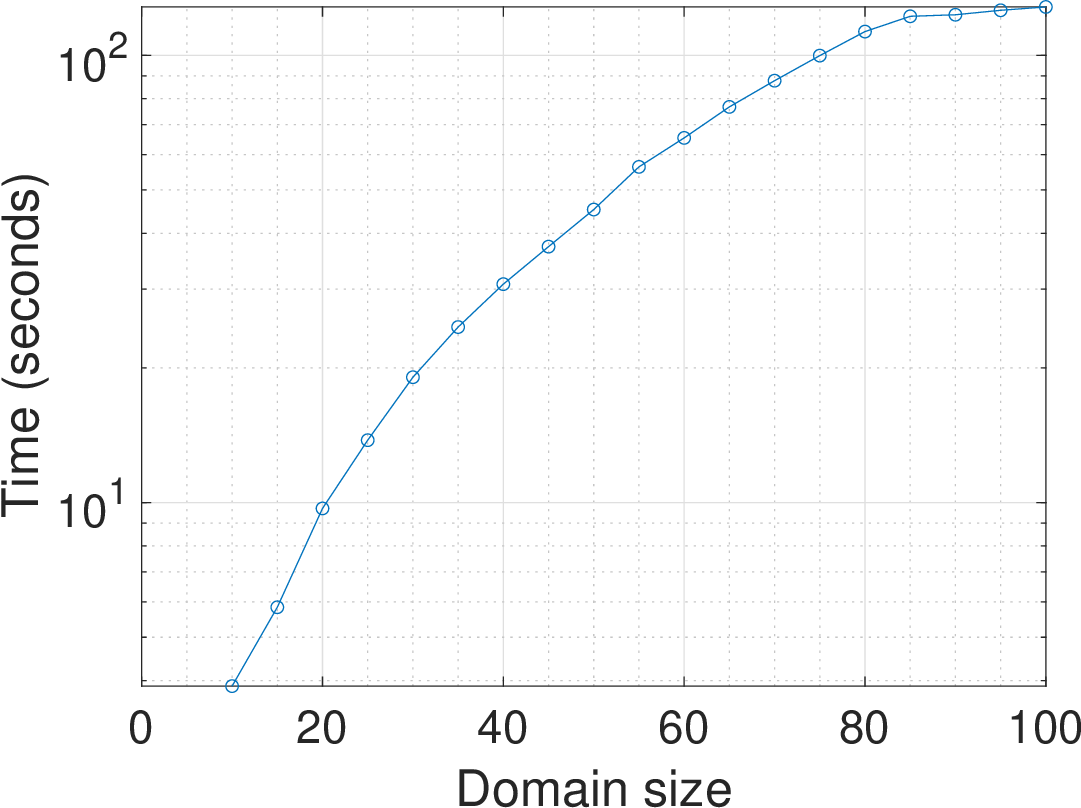}
            \caption{For optimizer method}
        \end{subfigure}
        \caption{Execution time for the initial guess method and $32$ iterations of the optimizer method}
        \label{fig:initGuessAndOptimizerTiming}
    \end{center}
\end{figure}

The execution time for initial guess and optimizer methods for varying domain sizes are shown in Figure \ref{fig:initGuessAndOptimizerTiming}. Python's built-in`time' module was used for this.
Observe that the initial guess method takes in the order of $10^{-3}$ seconds to execute, which is negligible compared to the execution time of the optimizer method, which is in the order of $10^{2}$ seconds.

\begin{figure}[h!]
    \begin{center}
    \includegraphics[width=0.22\textwidth]{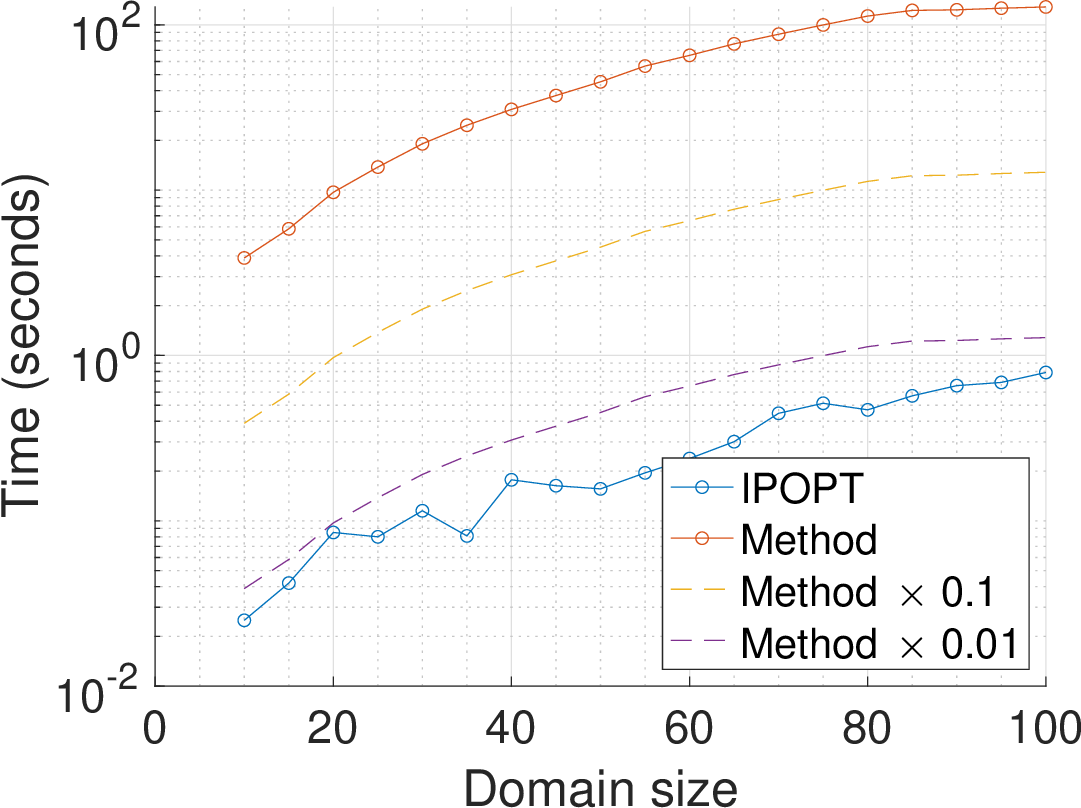}
    \caption{Comparing execution time for the proposed method with IPOPT}
    \label{fig:MethodVsIPOPTTiming}
    \end{center}
\end{figure}

A comparison between the execution time by the proposed method and IPOPT for $32$ iterations is shown in Figure \ref{fig:MethodVsIPOPTTiming}.
If implemented in a lower-level language, the method's execution time would be between $0.1$ and $0.01$ times the current. These are shown in the plot as well. Even the optimistic scaled version would be slower than IPOPT for the same number of iterations.

\subsubsection{Floating point operations}

\begin{figure}[h!]
    \begin{center}
        \begin{subfigure}[h]{0.22\textwidth}
            \includegraphics[width=\textwidth]{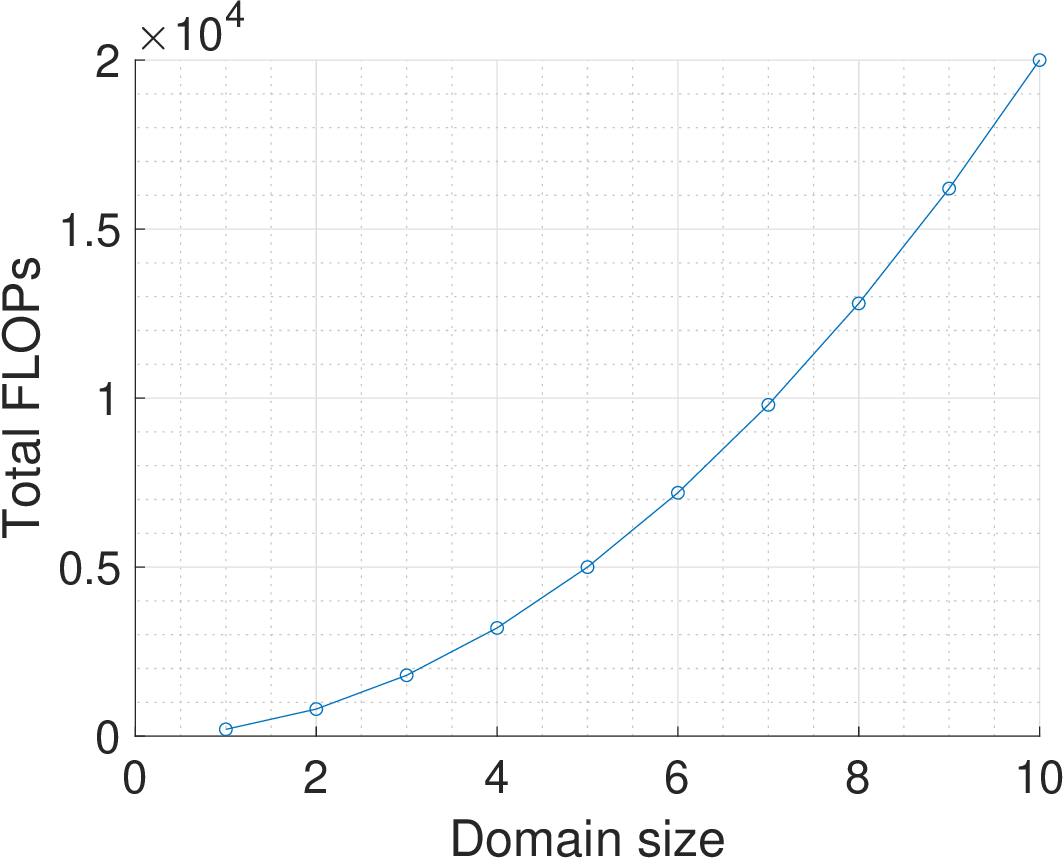}
            \caption{For initial guess method}
        \end{subfigure}
        \begin{subfigure}[h]{0.22\textwidth}
            \includegraphics[width=\textwidth]{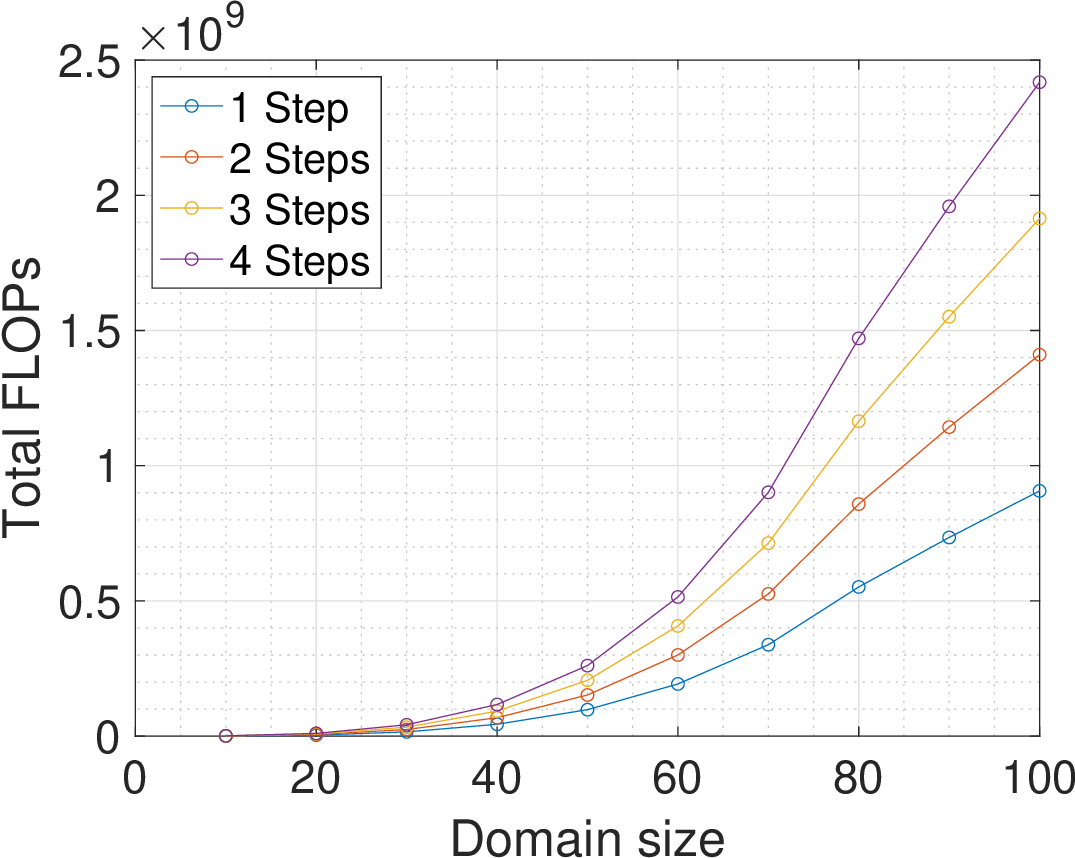}
            \caption{For optimizer method}
        \end{subfigure}
        \caption{FLOPs for the initial guess method and the optimizer method}
        \label{fig:FLOPsProposedMethod}
    \end{center}
\end{figure}

The number of floating point operations involved in the computation of the initial guess and the optimizer methods are shown in Figure \ref{fig:FLOPsProposedMethod}. This is calculated using PyTorch's profiler.
Observe a quadratic increase in the FLOPs for increasing domain size for both the initial guess and the optimizer methods. The number of FLOPs for the initial guess method is negligible compared to that for the optimizer method; the former is in the order of $10^4$ while the latter is in the order of $10^9$.

\begin{figure}[h!]
    \begin{center}
    \includegraphics[width=0.22\textwidth]{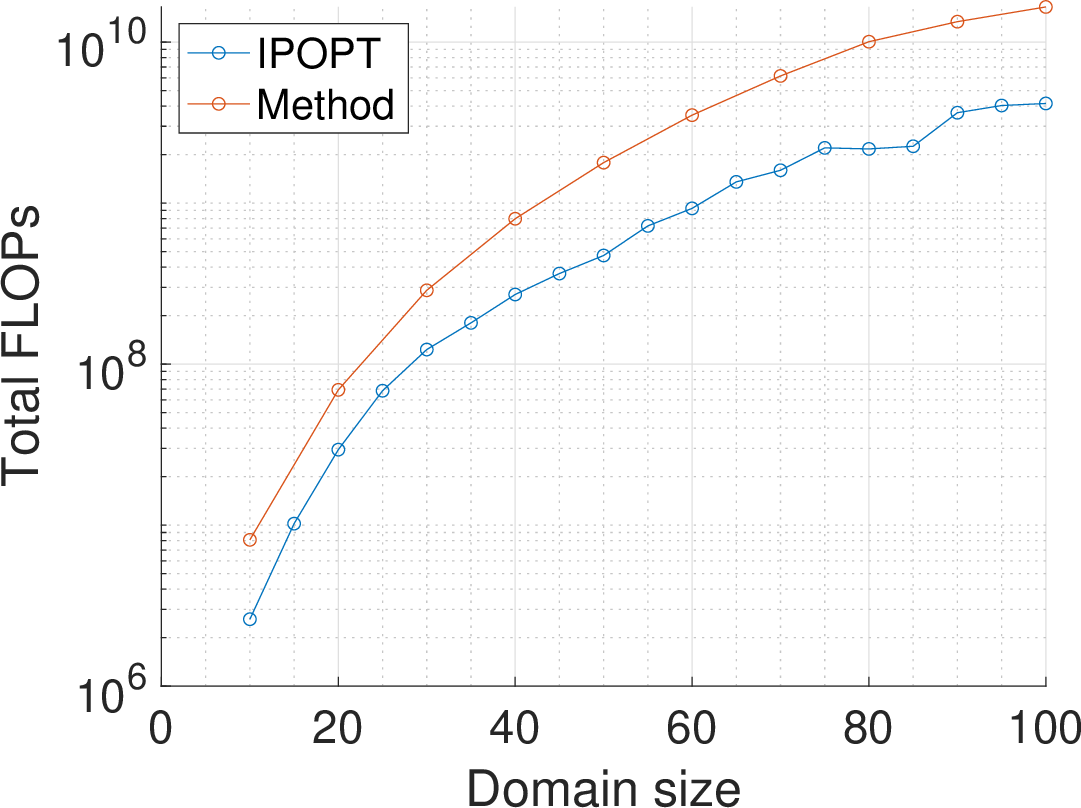}
    \caption{Comparing the number of FLOPs for the proposed method with that for IPOPT}
    \label{fig:MethodVsIPOPTFlops}
    \end{center}
\end{figure}

The comparison of the number of floating point operations between the proposed method and IPOPT for $32$ iterations is shown in Figure \ref{fig:MethodVsIPOPTFlops}.
Observe that the proposed method has slightly more floating point operations than IPOPT for the same number of iterations and hence, will likely be slower even if implemented in C++.

\subsection{Effect of larger domain sizes on accuracy}

\begin{figure}[h!]
    \begin{center}
        \begin{subfigure}[h]{0.22\textwidth}
            \includegraphics[width=\textwidth]{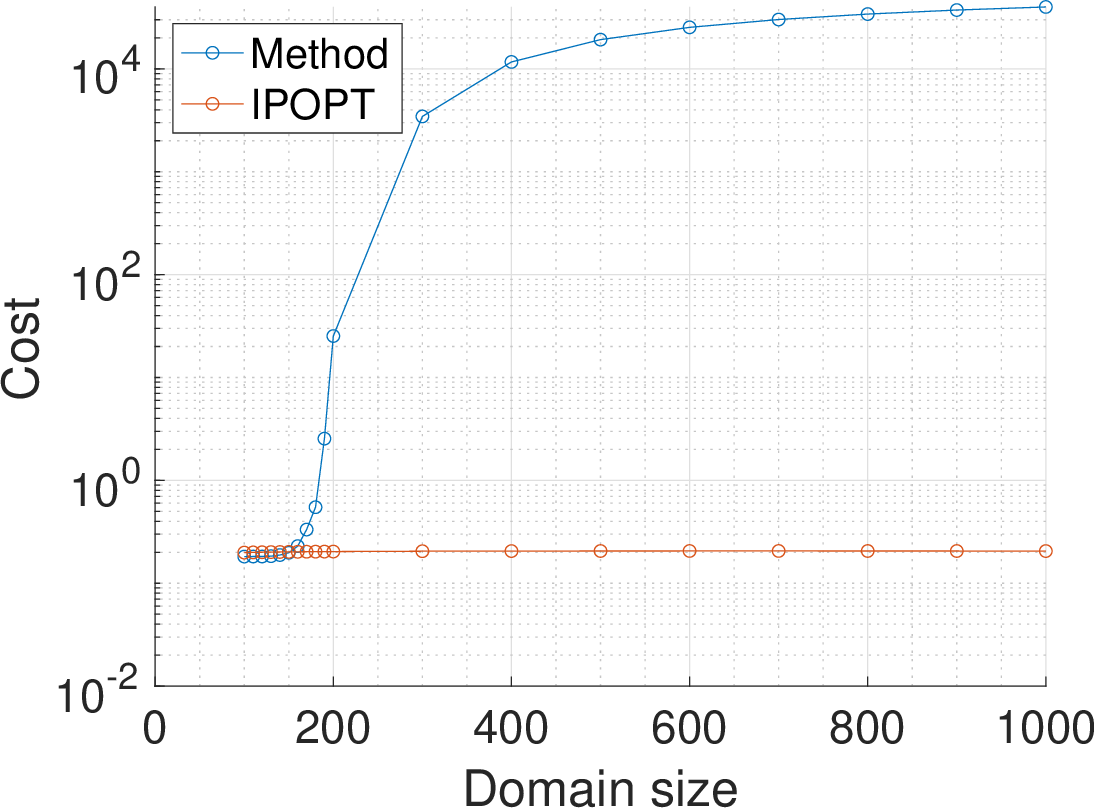}
            \caption{Problem index 0}
        \end{subfigure}
        \begin{subfigure}[h]{0.22\textwidth}
            \includegraphics[width=\textwidth]{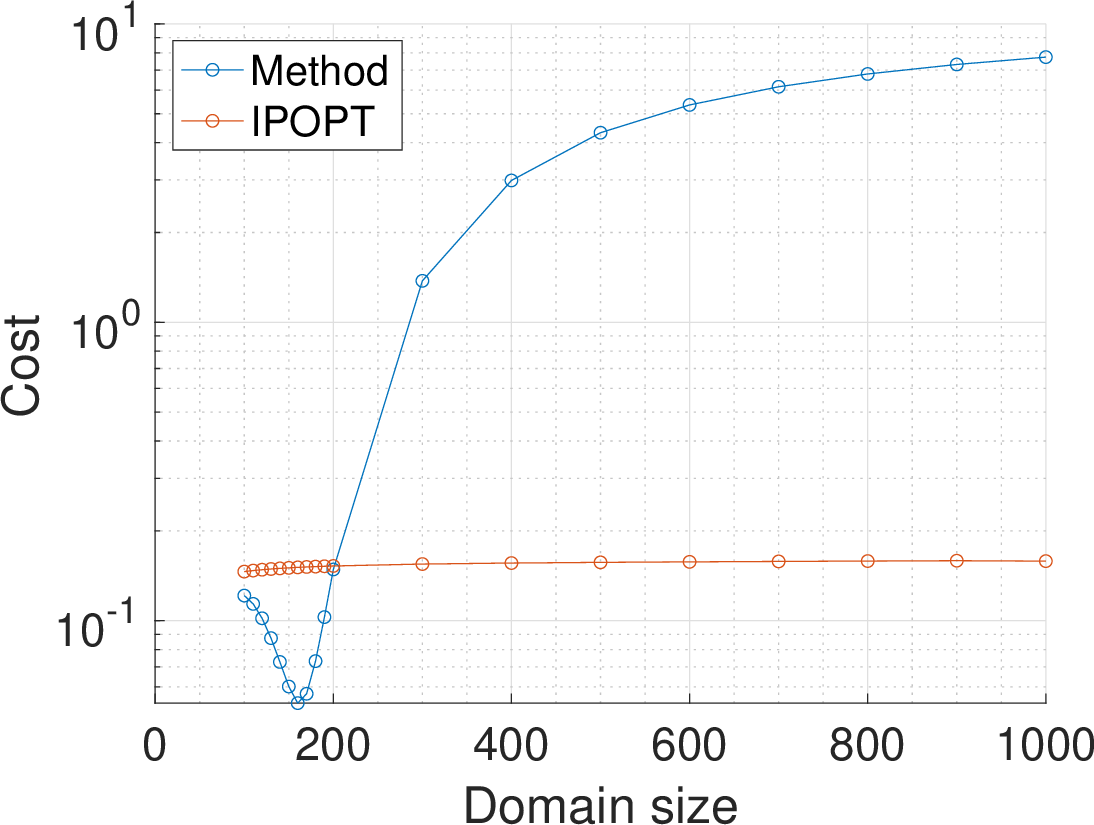}
            \caption{Problem index 17}
        \end{subfigure}
        \caption{Cost for larger domain sizes}
        \label{fig:ComparingCostForLargeN}
    \end{center}
\end{figure}

The data generated has domain sizes varying between $10$ and $100$. Figure \ref{fig:ComparingCostForLargeN} shows the cost calculated by the method and IPOPT for selected problems for domain sizes beyond this range.
Although the method may work till about a domain size of $150$, it fails miserably for larger domain sizes.

\section{Conclusions and discussions}

This study applies deep learning and reinforcement learning to boundary control problems by following an architecture framework, similar to most iterative optimization algorithms, containing two parts: one for making an initial guess, called the initial guess method, and the other, for making iterative improvement, called the optimizer method. The initial guess method is predominantly a convolution neural network. The optimizer method is a policy-gradient reinforcement learning method with the policy represented by a spatio-temporal neural network. Experimentation and analysis suggest that both initial guess and optimizer methods are effective. The initial guess method successfully outperforms trivial methods of generating an initial guess value for a given problem. The optimizer method achieves similar accuracy and performance to IPOPT, arriving at a lower cost solution in $50.97\%$ cases.

However, further research and exploration are required to provide additional conclusions and address the limitations of this research. These are summarized below.
\begin{itemize}
    \item Evaluating the method on real-world problems is in order. In future work, we aim to tackle problems from fluid mechanics, structural engineering and heat transfer domains by creating a dataset with real-world conditions and retraining the model to study its performance.
    \item Explore additional model extensions addressing the suboptimal performance on larger domain sizes and employ mitigation strategies.
    \item Strategies to reduce the computations per iteration and the overall number of iterations without affecting the accuracy need investigating. Some potential avenues are:
    \begin{enumerate}
        \item \textit{Faster PDE solvers}: Currently, the governing PDE is solved using the finite difference method. Improvements here could expedite the solver's performance. Neural network methods could perhaps be employed for this purpose as well.
        \item \textit{Faster gradient computation}: Currently, the gradients are computed using PyTorch's autograd. Explore ways to calculate or approximate it between the cost and the boundary values faster.
    \end{enumerate}
\end{itemize}

Overall, approaching boundary control problems with deep learning and reinforcement learning has merits. Further research may enable arriving at initial guess and optimizer methods that can achieve lower costs with fewer computations. The idea followed in this project may extend to optimization problems beyond boundary control problems; for instance, optimal power flow problems.

\bibliography{aaai23}
\pagebreak

\section{Appendix}
\section{Code and data}

The implementation code, generated dataset, results and analysis are available in \url{https://github.com/zenineasa/MasterThesis}.

To set up the Anaconda environment and IPOPT, follow the instructions in the Jupyter Notebook file named \textit{Code/setup.ipynb}. The environment information is available in the file named \textit{Code/environment.yml}. If the environment installation process does not work as specified in Jupyter Notebook, please manually install the packages listed in Table \ref{tab:softwareVersions}.

The program that generates synthetic data is in the file named \textit{Code/dataGenerator.py}, and the generated data is in the file named \textit{Code/Data/data.csv}. Files related to baseline generation and a few initial experiments are in the folder \textit{Code/experiment}. Files about applying deep learning and reinforcement learning methods to solve boundary control problems are in the folder \textit{Code/FinalAttempt}.

\section{Hardware and software setup}

The specification for the device used in training and validating all the neural networks is provided in Table \ref{tab:macSpec}. Testing was performed using several regular nodes in an HPC cluster with specifications listed in Table \ref{tab:icsCluster}. For performance evaluation, an old Linux machine with administrator privileges available to run a performance analyzing tool named 'perf' was used, the specification of which is described in Table \ref{tab:linuxSpec}. The information on the software and libraries used in this project are listed in Table \ref{tab:softwareVersions}.

\begin{table}[H]
    \centering
    \begin{tabular}{|p{0.12\textwidth}|p{0.30\textwidth}|}
        \hline
        \textbf{Component} & \textbf{Specification} \\
        \hline
        Hardware & MacBook Pro 2019, 16-inch \\
        Processor & 2.6 GHz Intel Core i7 \\
        No. of cores & 6 \\
        Graphics & AMD Radeon Pro 5300M 4 GB, \\ & Intel UHD Graphics 630 1536 MB \\
        Memory & 16 GB DDR4 @ 2667 MHz \\
        OS & macOS 13.3.1 \\
        \hline
    \end{tabular}
    \caption{Specification for the machine used for training, testing and validating neural networks}
    \label{tab:macSpec}
\end{table}
\begin{table}[H]
    \centering
    \begin{tabular}{|p{0.12\textwidth}|p{0.30\textwidth}|}
        \hline
        \textbf{Component} & \textbf{Specification} \\
        \hline
        Processor & Intel(R) Xeon(R) CPU E5-2650 v3 \\
        No. of cores & 20 \\
        Memory & 64GB DDR4 @ 2133MHz \\
        OS & CentOS Linux 8 (Core) \\
        \hline
    \end{tabular}
    \caption{Specification for the nodes used from the HPC Cluster}
    \label{tab:icsCluster}
\end{table}
\begin{table}[H]
    \centering
    \begin{tabular}{|p{0.12\textwidth}|p{0.30\textwidth}|}
        \hline
        \textbf{Component} & \textbf{Specification} \\
        \hline
        Hardware & Lenovo ideapad 330-15ARR \\
        Processor & AMD Ryzen 5 2500U \\
        No. of cores & 8 \\
        Graphics & AMD Radeon vega 8 graphics \\
        Memory & 8 GB ($2 \times$ 4 GB DDR4 @ 2400 MHz)\\
        OS & Ubuntu 22.04.02 LTS \\
        \hline
    \end{tabular}
    \caption{Specification of the machine used for performance evaluation}
    \label{tab:linuxSpec}
\end{table}
\begin{table}[H]
    \centering
    \begin{tabular}{|p{0.16\textwidth}|p{0.26\textwidth}|}
        \hline
        \textbf{Software / Library} & \textbf{Version} \\
        \hline
        Python & 3.9.16 \\
        PyTorch & 2.0.0 \\
        Pandas & 1.3.5 \\
        Anaconda & 4.13.0 \\
        IPOPT & 3.14.12 \\
        MATLAB & R2022b \\
        \hline
    \end{tabular}
    \caption{Versions of different software and libraries used}
    \label{tab:softwareVersions}
\end{table}

The implementation of the proposed method is written in Python using PyTorch and other libraries, while the analysis is in MATLAB, which makes it easier to know which files in the codebase are implementation and analysis related.

\section{Removing the effect of the sourcing term}

The following are two relationships related to the sourcing term of Laplace's equation.
\begin{itemize}
    \item {
        \textbf{Relationship between solutions for different sourcing term values}:
        Setting the boundary values to zero and numerically solving Poisson's equation for different sourcing term values, one can observe that the solution matrix when the sourcing term is $-20$ is double that of when it is $-10$. Similarly, for $-30$, $-40$ and $-50$, the solution matrices are three, four and five times that for $-10$, respectively.
    }
    \item {
        \textbf{Relationship between solutions for random and zero boundary values}:
        For a fixed domain size, if $A$ is the numerical solution for a given sourcing term with boundary values set to zero, $B$ is the numerical solution for random boundary values with the sourcing term set to zero, and $C$ is the numerical solution for the same random boundary values and the given sourcing term, then it can be observed that,
        \begin{equation}
            A + B = C
        \end{equation}
    }
\end{itemize}

Let's put these relationships together. Forward solving for Poisson's equation with a sourcing term of $-10$ using zero boundary values of dimensions $4 \times N$ will result in a matrix of size $N \times N$. For reusing, store these matrices for different values of $N$. To get the corresponding value for other sourcing terms, simply multiplying the values in the stored matrix for the desired size with the ratio of the required sourcing term and $-10$ would suffice. This matrix can be subtracted from the solution to Poisson's equation with a constant sourcing term to get the solution to Poisson's equation without a sourcing term!

\section{Initial guess method and edge values}

It was hard to conclude whether using the initial guess method or the edge values for the initial guess was more beneficial. Let us analyze the $3606$ cases where using the edge values for initial guess were observed to have lower cost and see if there is a pattern.

Partial correlation between the cost differences between the initial guess method and the edge values, and the problem parameters were calculated. A relatively high absolute partial correlation between the sourcing term and the cost differences was observed, which is a value of $0.4593$. Although this correlation in itself is not significantly high, because the correlation with every other parameter is below $0.07$, it may be useful to have a look into this.

\begin{figure}[h!]
    \begin{center}
    \includegraphics[width=0.22\textwidth]{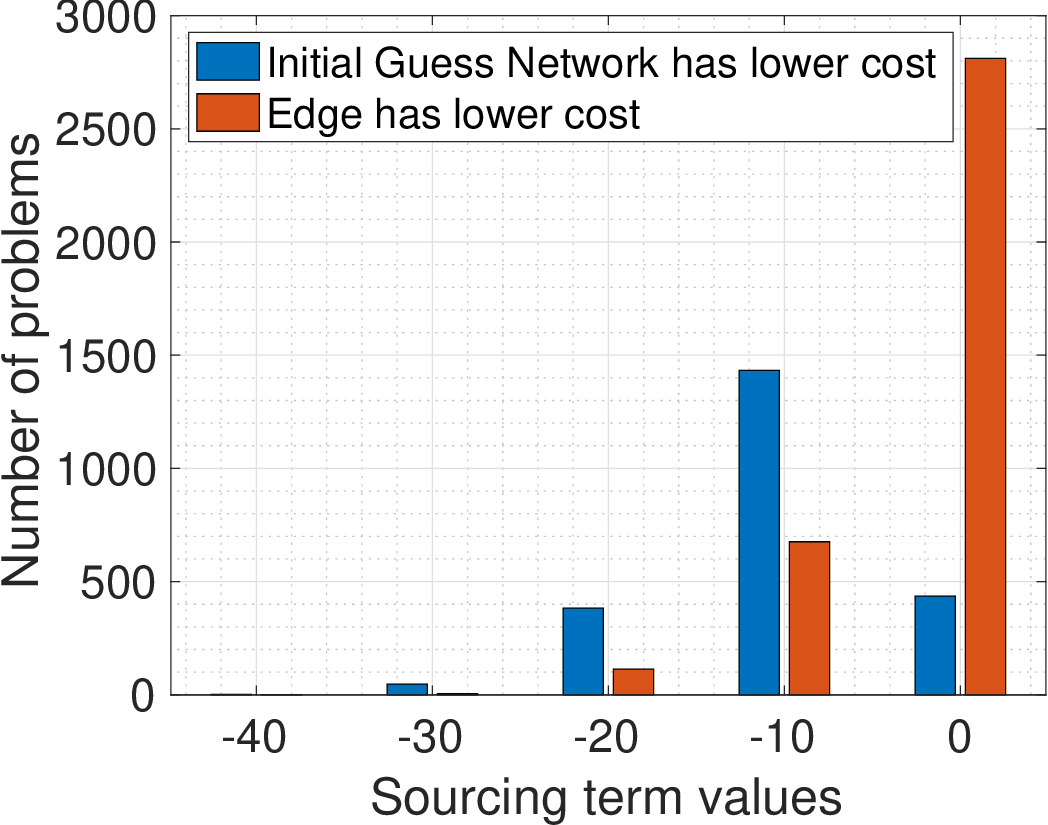}
    \caption{Number of problems for which initial guess method or edge values had lower cost for different sourcing term values}
    \label{fig:InitGuessNetworkVSEdge}
    \end{center}
\end{figure}

The number of cases where one method has a lower cost than the other can be compared easily. Figure \ref{fig:InitGuessNetworkVSEdge} shows a bar graph helping visualize the same. Observe that using the edge values might be more desirable when the sourcing value is $0$. In every other case, using the initial guess method would be more desirable.

This may indicate that one could create a better initial guess mechanism by adding a control statement that would let one guess the initial values as the edge values if the sourcing term value is zero and use the initial guess method in every other case. However, it is also likely that there exists some bias in the problem generator that has caused this.

\section{Adam and RMSProp in optimizer method}

There are two reasons why Adam and RMSProp were included along with the spatio-temporal network in the optimizer method.
\begin{enumerate}
    \item Adam and RMSProp have been widely used in deep learning as optimization algorithms to tune network weights. They have demonstrated to work well in quite a lot of scenarios.
    \item Adam and RMSProp appear to be guiding the spatio-temporal network to train. Without them, the best values observed would be the first set of values (the output of the initial guess method), and the network iterations would not be considered during backpropagation.
\end{enumerate}

Indeed, the second issue could have been addressed by tweaking the training method. It is also true that either one of Adam or RMSProp could have been used for this and not both simultaneously. Furthermore, several other optimizers are being used in research like Adaptive Gradient Algorithm (Adagrad), AdaDelta, Stochastic Gradient Descend (SGD), etc. This is just the decision that was made.

\subsection{Contribution of Adam, RMSProp and spatio-temporal parts}

The learnt learning rates for Adam, RMSProp and the Spatio-temporal part of the optimizer are respectively $0.0223$, $0.0221$ and $0.0645$. However, this does not conclusively mean that the contribution of Spatio-temporal is higher than that by Adam and RMSProp.

The information regarding contribution of each of the three parts are recorded for $32$ optimizer iterations along with the domain size and order of iteration. The average magnitude of contribution per optimizer iteration for Adam, RMSProp and the Spatio-temporal part are, respectively, $0.0151$, $0.0133$ and $0.0251$. Clearly, the contribution from the network is higher in magnitude.

\begin{figure}[h!]
    \begin{center}
    \includegraphics[width=0.22\textwidth]{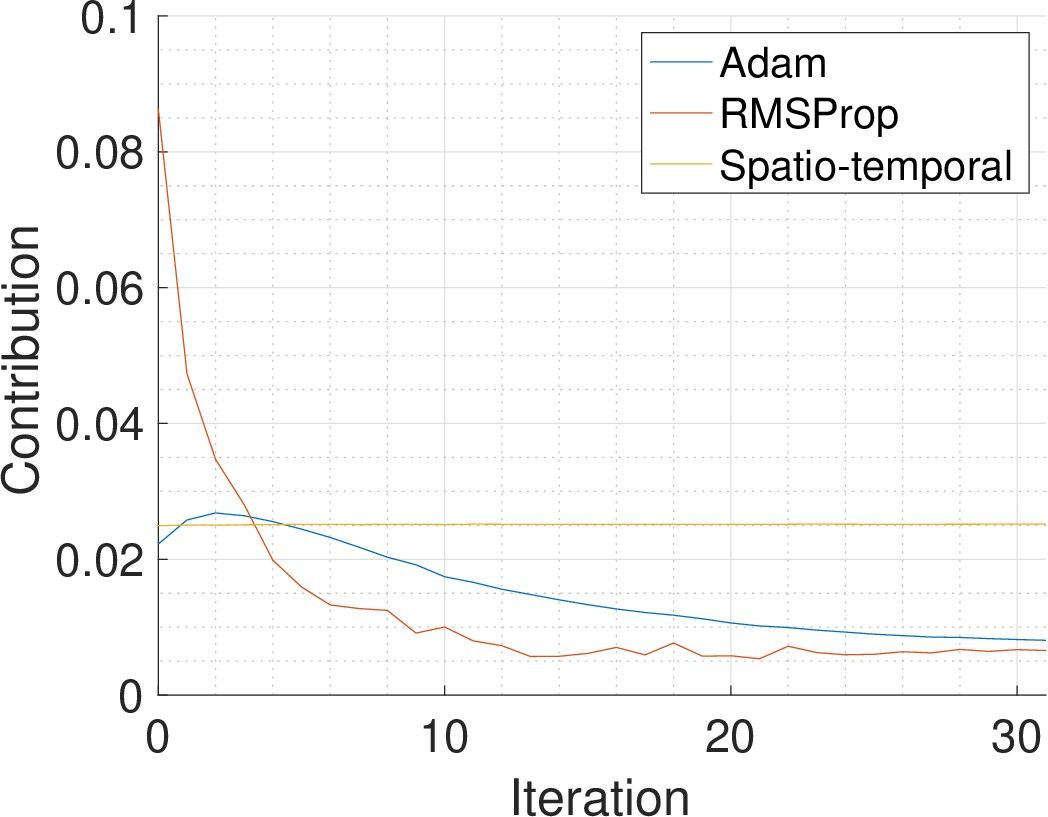}
    \caption{Comparing the contribution by Adam, RMSProp and Spatio-temporal part over different iteration}
    \label{fig:contributionByAdamRMSPropAndNet}
    \end{center}
\end{figure}

Furthermore, the partial correlation between Adam, RMSProp and Spatio-temporal parts with the number of iterations are $-0.6017$, $-0.3360$ and $-0.0086$, respectively. The effects of this correlation are visualized in Figure \ref{fig:contributionByAdamRMSPropAndNet}. For Adam and RMSProp, the values are as expected; as the number of iterations increases, the contribution would decrease, indicating that they are converging to a solution.
However, for some peculiar reason, the magnitude of contribution from the Spatio-temporal part does not seem to change, suggesting that this part may not be doing anything meaningful, as it may just be outputting some constant value. Let us investigate whether that is true or not.

\subsubsection{Investigating Spatio-temporal part}

This investigation is to know if the network is doing something meaningful and not just outputting the biasses. If it is all about the bias terms, then for different inputs to the Spatio-temporal part, the output would more or less be a constant value.

In the setup, a minor modification is made to store the value of the output of the Spatio-temporal part in the matrix form for the first iteration. Then, for every subsequent iteration, the output of the Spatio-temporal part at that iteration is subtracted with the stored value, following which the mean of the absolute values are calculated and logged.
\begin{figure}[h!]
    \begin{center}
        \begin{subfigure}[h]{0.22\textwidth}
            \includegraphics[width=\textwidth]{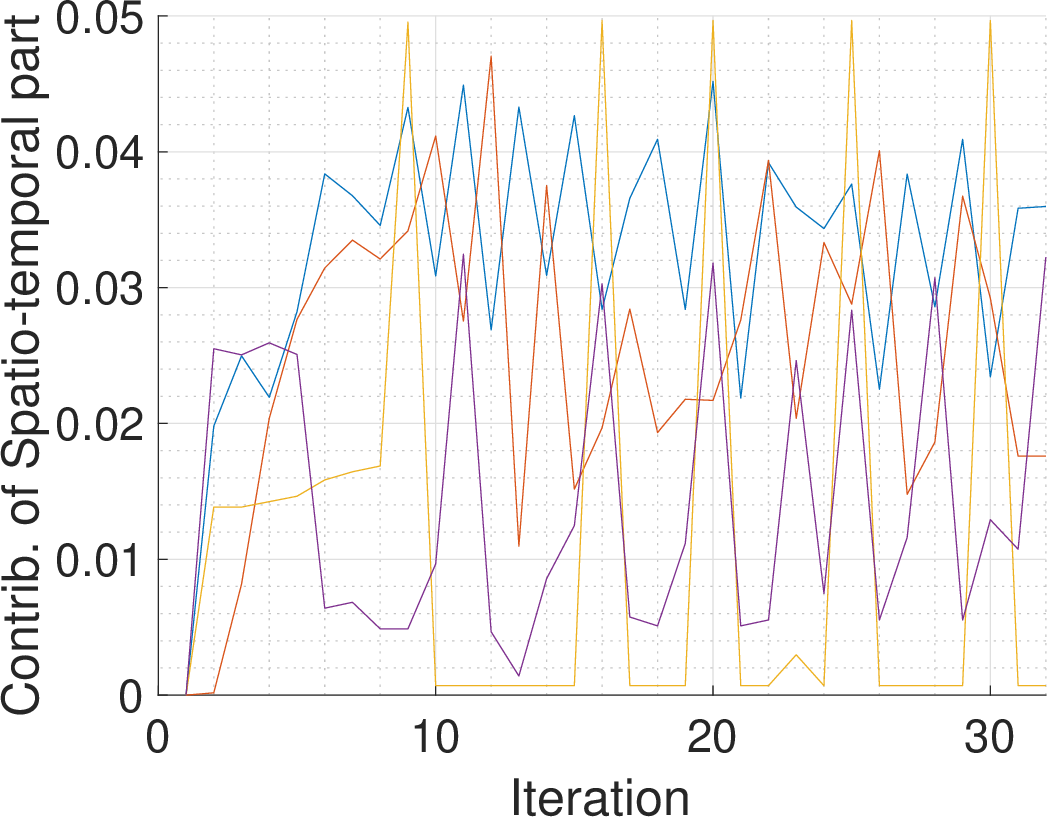}
            \caption{For four different problems}
        \end{subfigure}
        \begin{subfigure}[h]{0.22\textwidth}
            \includegraphics[width=\textwidth]{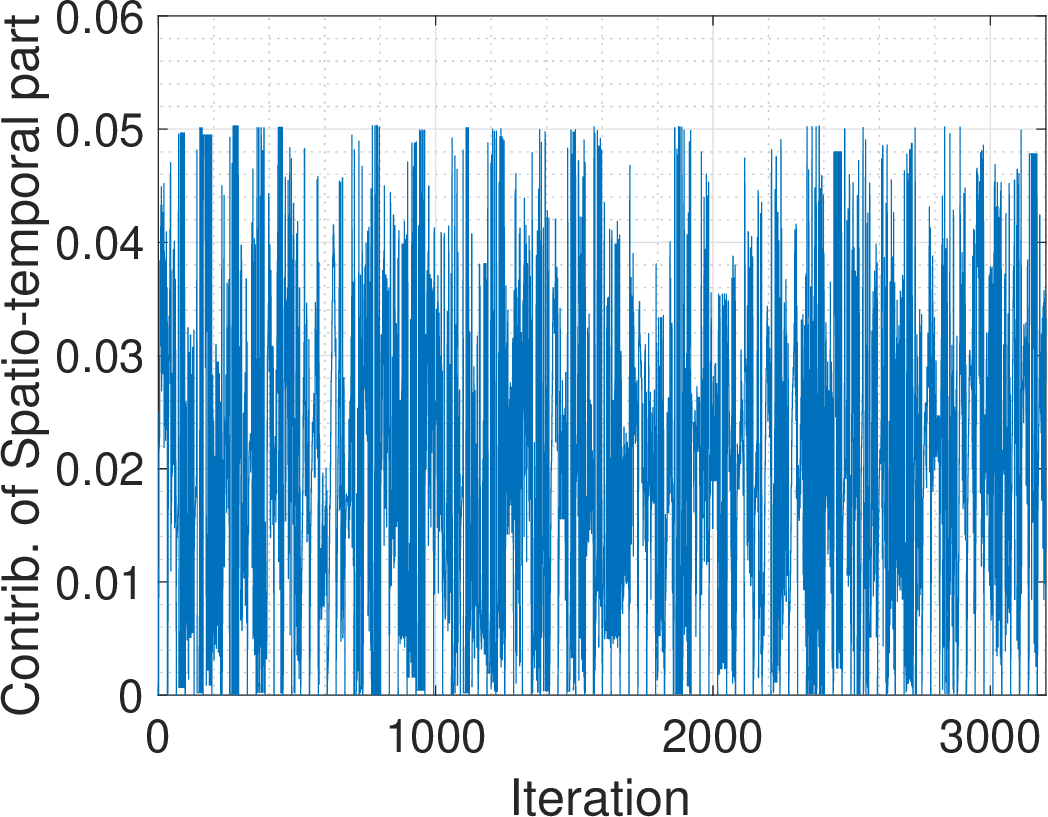}
            \caption{For 3200 iterations from different problems}
        \end{subfigure}
        \caption{Contribution by Spatio-temporal part visualized}
        \label{fig:contributionByNet}
    \end{center}
\end{figure}
Figure \ref{fig:contributionByNet} helps visualize logged information. Observe that the contribution from the network is not constant. The variations observed are between $0$ and $0.0503$, which is significant.

Although it is still quite interesting to see that the contribution by the Spatio-temporal network has a constant-ish mean absolute value, it does not mean in any way that the network itself is giving us a constant-ish value.

\end{document}